# Safe Exploration of State and Action Spaces in Reinforcement Learning

**Javier García**　　　　　　　　　　　　　　　　FJGPOLO@INF.UC3M.ES
**Fernando Fernández**　　　　　　　　　　　　　FFERNAND@INF.UC3M.ES
*Universidad Carlos III de Madrid,*
*Avenida de la Universidad 30,*
*28911 Leganés, Madrid, Spain*

## Abstract

In this paper, we consider the important problem of safe exploration in reinforcement learning. While reinforcement learning is well-suited to domains with complex transition dynamics and high-dimensional state-action spaces, an additional challenge is posed by the need for safe and efficient exploration. Traditional exploration techniques are not particularly useful for solving dangerous tasks, where the trial and error process may lead to the selection of actions whose execution in some states may result in damage to the learning system (or any other system). Consequently, when an agent begins an interaction with a dangerous and high-dimensional state-action space, an important question arises; namely, that of how to avoid (or at least minimize) damage caused by the exploration of the state-action space. We introduce the PI-SRL algorithm which safely improves suboptimal albeit robust behaviors for continuous state and action control tasks and which efficiently learns from the experience gained from the environment. We evaluate the proposed method in four complex tasks: automatic car parking, pole-balancing, helicopter hovering, and business management.

## 1. Introduction

Reinforcement learning (RL) (Sutton & Barto, 1998) is a type of machine learning whose main goal is that of finding a policy that moves an agent optimally in an environment, generally formulated as a **M**arkov **D**ecision **P**rocess (MDP). Many RL methods are being used in important and complex tasks (e.g., robot control see Smart & Kaelbling, 2002; Hester, Quinlan, & Stone, 2011, stochastic games see Mannor, 2004; Konen & Bartz-Beielstein, 2009 and control optimization of complex dynamical systems see Salkham, Cunningham, Garg, & Cahill, 2008). While most RL tasks are focused on maximizing a long-term cumulative reward, RL researchers are paying increasing attention not only to long-term reward maximization, but also to the safety of approaches to **S**equential **D**ecision **P**roblems (SDPs) (Mihatsch & Neuneier, 2002; Hans, Schneegass, Schäfer, & Udluft, 2008; Martín H. & Lope, 2009; Koppejan & Whiteson, 2011). Well-written reviews of these matters can also be found (Geibel & Wysotzki, 2005; Defourny, Ernst, & Wehenkel, 2008). Nevertheless, while it is important to ensure reasonable system performance and consider the safety of the agent (e.g., avoiding collisions, crashes, etc.) in the application of RL to dangerous tasks, most exploration techniques in RL offer no guarantees on both issues. Thus, when using RL techniques in dangerous control tasks, an important question arises; namely, how can we ensure that the exploration of the state-action space will not cause damage or injury while, at the same time, learning (near-)optimal policies? The matter, in other words, is one of ensuring that the agent be able to explore a dangerous environment both safely and efficiently. There are many domains where the exploration/exploitation process may lead to catastrophic states or actions for the learning agent (Geibel & Wysotzki, 2005). The helicopter hovering control task is one such case involving high risk, since some policies can crash the helicopter, incurring catastrophic negative reward. Exploration/exploitation strategies such as $\epsilon - greedy$ may even result in constant helicopter crashes (especially where there is a high probability of random action selection). Another example can be found in portfolio theory where analysts are expected to find a portfolio that maximizes profit while avoiding risks of considerable losses (Luenberger, 1998). Since the maximization of expected returns does not necessarily prevent rare occurrences of large negative outcomes, a different criteria for safe exploration is needed. The exploration process in which new policies are evaluated must be conducted with extreme care. Indeed, for such environments, a method is required which not only explores the state-action space, but which does so in a safe manner.

In this paper, we propose the **P**olicy **I**mprovement through **S**afe **R**einforcement **L**earning (PI-SRL) algorithm for safe exploration in dangerous and continuous control tasks. Such a method requires a predefined (and safe) baseline policy which is assumed to be suboptimal (otherwise, learning would be pointless). Predefined baseline policies have been used in different ways by other approaches. In the work of Koppejan and Whiteson (2011), single-layers perceptrons are evolved, albeit starting from a prototype network whose weights correspond to a baseline policy provided by helicopter control task competition software (Abbeel, Coates, Hunter, & Ng, 2008). This approach can be viewed as a simple form of population seeding which has proven to be advantageous in numerous evolutionary methods (e.g. see Hernández-Díaz, Coello, Perez, Caballero, Luque, & Santana-Quintero, 2008; Poli & Cagnoni, 1997). In the work of Martín and de Lope (2009), the weights of neural networks are also evolved by inserting several baseline policies (including that provided in the helicopter control task competition software) into the initial population. To minimize the possibility of evaluating unsafe policies, their approach prevents crossover and mutation operators from permitting anything more than tiny changes to the initial baseline policies. In this paper, we present the PI-SRL algorithm, a novel approach to improving baseline policies in dangerous domains using RL. The PI-SRL algorithm is composed of two different steps. In the first, baseline behavior (robust albeit suboptimal) is approximated using behavioral cloning techniques (Anderson, Draper, & Peterson, 2000; Abbott, 2008). In order to achieve this goal, case-based reasoning (CBR) techniques (Aamodt & Plaza, 1994; Bartsch-Sprl, Lenz, & Hbner, 1999) were used which have been successfully applied to imitation tasks in the past (Floyd & Esfandiari, 2010; Floyd, Esfandiari, & Lam, 2008). In the second step, the PI-SRL algorithm attempts to safely explore the state-action space in order to build a more accurate policy from previously-learned behavior. Thus, the set of cases (i.e., state-action pairs) obtained in the previous phase is improved through the safe exploration of the state-action space. To perform this exploration, small amounts of Gaussian noise are randomly added to the greedy actions of the baseline policy approach. The exploration strategy has been used successfully in previous works (Argall, Chernova, Veloso, & Browning, 2009; Van Hasselt & Wiering, 2007).

The novelty of the present study is in the use of two new, main components: (i) a *risk function* to determine the degree of risk of a particular state and (ii) a *baseline behavior* capable of producing safe actions in supposedly risky states (i.e., states that can lead to damage or injury). In addition, we present a new definition of *risk* based on what for the agent is *unknown* and *known space*. As will be described in Section 5 in greater detail, this new definition is completely different from traditional definitions of "risk" found in the literature (Geibel, 2001; Mihatsch & Neuneier, 2002; Geibel & Wysotzki, 2005). The paper also reports the experimental results obtained from the application of the new approach in four different domains: (i) automatic car parking (Lee & Lee, 2008), (ii) pole-balancing (Sutton & Barto, 1998), (iii) 2009 RL Competition helicopter hovering (Ng, Kim, Jordan, & Sastry, 2003) and (iv) business management (Borrajo, Bueno, de Pablo, Santos, Fernández, García, & Sagredo, 2010). In each domain, we propose the learning of a near-optimal policy which, in the learning phase, will minimize car crashes, pole disequilibrium, helicopter crashes and company bankruptcies, respectively. It is important to note that the comparison of our approach with an agent with an optimal exploration policy is not possible since, in the proposed domains (each with a high-dimensional and continuous state and action space, as well as complex stochastic dynamics), we do not know what the optimal exploration policy is.

Regarding the organization of the remainder of the paper, Section 2 introduces key definitions, while Section 3 describes in detail the learning approach proposed. In Section 4, the evaluation performed in the four above mentioned domains is presented. Section 5 discusses related work and Section 6 summarizes the main conclusions of our study. In these sections, the term *return* is used to refer to the expected cumulative future discounted reward $R = \sum_{t=0}^{\infty} \gamma^t r_t$, while the term *reward* is used to refer to a single real value used to evaluate the selection of an action in a particular state and it is denoted by $r$.

## 2. Definitions

To illustrate the concept of safety used in our approach, a navigation problem is presented below in Figure 1. In the navigation problem presented in Figure 1, a control policy must be learned to get from a particular start state to a goal state, given a set of demonstration trajectories. In this environment, we assume the task to be difficult due to a stochastic and complex dynamic of the environment (e.g., an extremely irregular surface in the case of a robot navigation domain or wind effects in the case of the helicopter hover task). This stochasticity makes it impossible to complete the task using exactly the same trajectory every time. Additionally, the problem supposes that a set of demonstrations from a baseline controller performing the task (the continuous black lines) are also given. This set of demonstrations is composed of different trajectories covering a well-defined region of the state space (the region within the rectangle).

Our approach is based on the addition of small amounts of Gaussian noise or perturbations to the baseline trajectories in order to find new and better ways of completing the task. This noise will affect the baseline trajectories in different ways, depending on the amount of noise added which, in turn, depends on the amount of risk to be taken. If no risk is desired, the noise added to the baseline trajectories will be 0 and, consequently, no new or improved behavior will be discovered (nevertheless, the robot will never fall off the cliff and the helicopter will never crash). If, however, an intermediate level of risk is desired, small amounts of noise will be added to the baseline trajectories and new trajectories (the dotted blue lines) to complete the task are discovered. In some cases, the exploration of new trajectories leads the robot to unknown regions of the state space (the dashed red lines). The robot is assumed to be able to detect such situations with a risk function and use the baseline behavior to return to safe, known states. If, instead, a very high risk is desired, large amounts of noise will be added to the baseline trajectories, leading to the discovery of new trajectories (but also to a higher probability that the robot gets damaged). The iteration of this process leads the robot to progressively and safely explore the state and action spaces in order to find new and improved ways to complete the task. The degree of safety in the exploration, however, will depend on the risk taken.

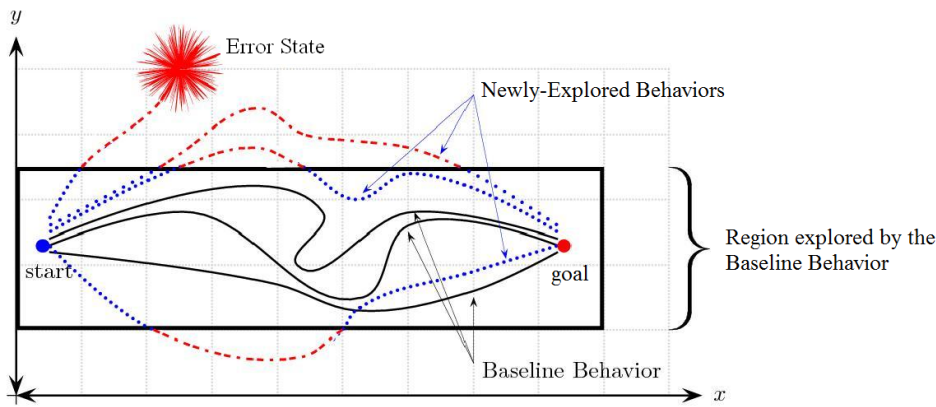

Figure 1: Exploration strategy based on the addition of small amounts of noise to baseline policy behavior. Continuous lines represent the baseline behavior, while newly explored behaviors are indicated by the dotted and dashed lines.

## 2.1 Error and Non-Error States

In this paper, we follow as far we can the notation presented in Geibel et al. (2005) for the definition of our concept of *risk*. In their study, Geibel et al. associate risk with *error states* and *non-error states*, with the former understood as a state in which it is considered undesirable or dangerous to enter.

**Definition 1** *Error and non-error states. Let $S$ be a set of states and $\Phi \subset S$ the set of error states. A state $s \in \Phi$ is an undesirable terminal state where the control of the agent ends when $s$ is reached with damage or injury to the agent, the learning system or any external entities. The set $\Gamma \subset S$ is considered a set of non-error terminal states with $\Gamma \cap \Phi = \emptyset$ and where the control of the agent ends normally without damage or injury.*

In terms of RL, if the agent enters an error state, the current episode ends with damage to the learning system (or other systems); whereas if it enters a non-error state, the episode ends normally and without damage. Thus, Geibel et al. define the risk of $s$ with respect to policy $\pi$, $\rho^{\pi}(s)$, as the probability that the state sequence $(s_i)_{i \geq 0}$ with $s_0 = s$, generated by the execution of policy $\pi$, terminates in an error state $s' \in \Phi$. By definition, $\rho^{\pi}(s) = 1$ if $s \in \Phi$. If $s \in \Gamma$, then $\rho^{\pi}(s) = 0$ because $\Phi \cap \Gamma = \emptyset$. For states $s \notin \Phi \cup \Gamma$, the risk taken depends on the actions selected by the policy $\pi$. With these definitions, we have the theoretical framework with which to introduce our own definition of the risk associated with *known* and *unknown states.*

## 2.2 Known and Unknown States in Continuous Action and State Spaces

We assume a continuous, $n$-dimensional state space $S \subset \Re^n$ where each state $s = (s_1, s_2, \ldots, s_n) \in S$ is a vector of real numbers and each dimension has an individual domain $D_i^s \subset \Re$. Similarly, we assume a continuous and $m$-dimensional action space $A \subset \Re^m$ where each action $a = (a_1, a_2, \ldots, a_m) \in A$ is a vector of real numbers and each dimension has an individual domain $D_i^a \subset \Re$. Additionally, the agent considered here is endowed with a memory, or case-base $B$, of the size $\eta$. Each memory element represents a state-action pair, or case, the agent has experienced before.

**Definition 2 *(Case-base).* *A case-base is a set of cases $B = \{c_1 \ldots, c_\eta\}$. Every case $c_i$ consists of a state-action pair $(s_i, a_i)$ the agent has experienced in the past and with an associated value $V(s_i)$. Thus, $c_i = <s_i, a_i, V(s_i)>$, where the first element represents the case's problem part and corresponds to the state $s_i$, the following element $a_i$ depicts the case solution (i.e., the action expected when the agent is in the state $s_i$) and the final element $V(s_i)$ is the value function associated with the state $s_i$. Each state $s_i$ is composed of $n$ continuous state variables and each action $a_i$ is composed of $m$ continuous action variables.*

When the agent receives a new state $s_q$, it first retrieves the nearest neighbor of $s_q$ in $B$ according to a given similarity metric and then performs the associated action. In this paper, we use Euclidean distance as our similarity metric (Equation 1).

$$d(s_q, s_i) = \sqrt{\sum_{j=0}^{n} (s_{q,j} - s_{i,j})^2} \tag{1}$$

The Euclidean distance metric is useful when the value function is expected to be continuous and smooth throughout the state space (Santamaría, Sutton, & Ram, 1998). However, since the value function is unknown a priori and the Euclidean distance metric is not particularly suitable for many problems, many researchers have begun to ask how the distance metric itself can learn or adapt in order to achieve better results (Taylor, Kulis, & Sha, 2011). While the use of distance metric learning techniques would certainly be desirable in order to induce a more powerful distance metric for a specific domain, such a consideration lies outside the scope of the present study. In this paper, therefore, we have focused only on domains in which Euclidean distance has been proven successful (i.e., it has been successfully applied to car parking (Cichosz, 1995), pole-balancing (Martin H & de Lope, 2009), helicopter hovering control (Martin H & de Lope, 2009) and SIMBA (Borrajo et al., 2010).

Traditionally, case-based approaches use a *density threshold* $\theta$ in order to determine when a new case should be added to the memory. When the distance of the nearest neighbor to $s_q$ is greater than $\theta$, a new case is added. In this sense, the parameter $\theta$ defines the size of the classification region for each case in $B$ (Figure 2). If a new case $s_q$ is within the classification region of a case $c_i$, it is considered to be a known state. Hence, the cases in $B$ describe a case-based policy of the agent $\pi_B^\theta$ and its associated value function $V^{\pi_B^\theta}$.

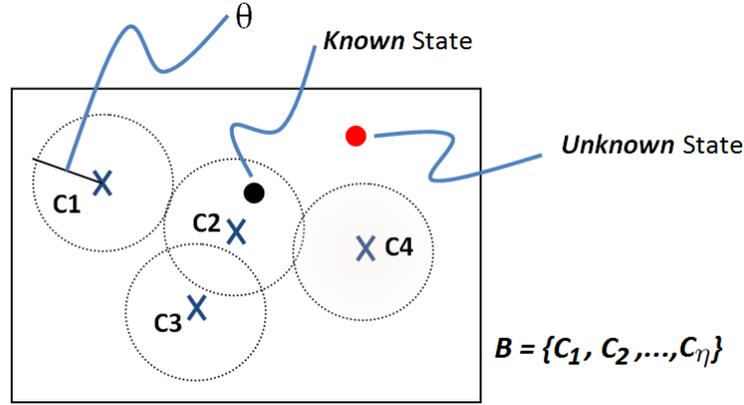

Figure 2: Known and Unknown states.

**Definition 3 (*Known/Unknown states*).** *Given a case-base $B = \{c_1 \dots, c_\eta\}$ composed of cases $c_i = (s_i, a_i, V(s_i))$ and a density threshold $\theta$, a state $s_q$ is considered known when $\min_{1 \leq i \leq \eta} d(s_q, s_i) \leq \theta$ and unknown in all other cases. Formally, $\Omega \subseteq S$ is the set of known states, while $\Upsilon \subseteq S$ is the set of unknown states with $\Omega \cap \Upsilon = \emptyset$ and $\Omega \cup \Upsilon = S$.*

With Definition 3, states can be identified as known or unknown. When the agent receives a new state $s \in \Omega$, it performs the action $a_i$ of the case $c_i$ for which $d(s, s_i) = \min_{1 \leq j \leq \eta} d(s, s_j)$. However, if the agent receives a state $s \in \Upsilon$ where, by definition, the distance to any state in $B$ is larger than $\theta$, no case is retrieved. Consequently, the action to be performed from that state is unknown to the agent.

**Definition 4 (*Case-Based risk function*).** *Given a case base $B = \{c_1 \dots, c_\eta\}$ composed of cases $c_i = (s_i, a_i, V(s_i))$, the risk for each state $s$ is defined as Equation 2.*

$$\varrho^{\pi_B^\theta}(s) = \begin{cases} 0 & \text{if } \min_{1 \leq j \leq \eta} d(s, s_j) < \theta \\ 1 & \text{otherwise} \end{cases} \qquad (2)$$

Thus, $\varrho^{\pi_B^\theta}(s) = 1$ holds if $s \in \Upsilon$ (i.e., $s$ is unknown), such that the state $s$ is not associated with any case and, hence, the action to be performed in the given situation is unknown. If $s \in \Omega$, then $\varrho^{\pi_B^\theta}(s) = 0$.

**Definition 5 (*Safe case-based policy*).** *The case-based policy $\pi_B^\theta$ derived from a case-base $B = \{c_1 \dots, c_\eta\}$ is safe when, from any initial known state $s_0$ with respect to $B$, the execution of $\pi_B^\theta$ always produces known non-error states with respect to $B$.*

$$\forall s_0 \mid \varrho^{\pi_B^\theta}(s_0) = 0, \ then \ \forall (s_i)_{i>0}^{\pi_B^\theta} \ \varrho^{\pi_B^\theta}(s_i) = 0 \qquad (3)$$

Additionally, it is assumed here that the probability that the state sequence $(s_i)_{i \geq 0}$ from any known state $s_0 \in \Omega$, generated by executing policy $\pi_B^\theta$, terminates in an error state $s \in \Phi$ is $\rho^{\pi_B^\theta}(s_0) = 0$ (i.e., $\Omega \cap \Phi = \emptyset$).

**Definition 6** (*Safe case-based coverage*). *The coverage of a single state $s$ with respect to a safe case-base $B = \{c_1 \ldots, c_\eta\}$ is defined as the state $s_i$ for which $\min_{1 \le i \le \eta} d(s, s_i) \le \theta$. Therefore, we assume that the safe case-based does not provide actions for the entire state space, but rather only for known states $s \in \Omega$.*

Figure 3 graphically represents the relationship between known/unknown and error/non-error states. The green area in the image denotes the safe case-based policy $\pi_B^\theta$ learnt, an area of the state space corresponding to the initial known space. An agent following the policy $\pi_B^\theta$ will always be in the green area and all resulting episodes will end without damages. Consequently, a subset of non-error states will also form part of the known space. Formally, let $\Gamma_\Omega$ and $\Gamma_\Upsilon$ be subsets of non-error states belonging to the known and unknown spaces, respectively, with $\Gamma_\Omega \cup \Gamma_\Upsilon = \Gamma$. Then $\Gamma_\Omega \subset \Omega$. The yellow area in the Figure, by contrast, represents the unknown space $\Upsilon$. In this space will be found all error states, as well as a subset of remaining non-error states. Formally, $\Gamma_\Upsilon \subset \Upsilon$ and $\Phi \subset \Upsilon$.

Understood in this way, the PI-SRL algorithm can be summed up as follows:

- As a first step, learn the known space (green area) from the safe case-based policy $\pi_B^\theta$.

- As a second step, adjust the known space (green area) and unknown space (yellow area) in order to explore new and improved behaviors while avoiding error states (red area). During this process of adjusting the known space to the space used for safe and better policies, the algorithm can "forget" ineffectual known states, as will be shown in Section 4.

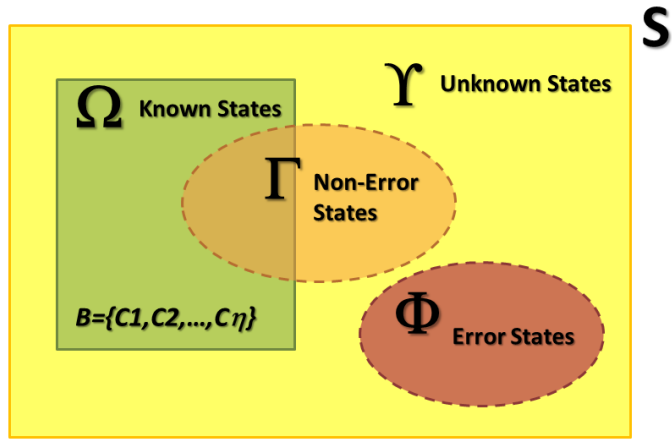

Figure 3: Known/unknown and error/non-error states given the Case Base $B$.

## 2.3 The Advantages of Using Prior Knowledge and Predetermined Exploration Policies

In the present subsection, the advantages of using teacher knowledge in RL, namely (i) to provide initial knowledge about the task to be learned and (ii) to support the exploration process, are highlighted. Furthermore, we explain why we believe this knowledge to be indispensable in RL for tackling highly complex and realistic problems with large, continuous state and action spaces and in which a particular action may result in an undesirable consequence.

### 2.3.1 Providing Initial Knowledge about the Task

Most RL algorithms begin learning without any previous knowledge about the task to be learnt. In such cases, exploration strategies such as $\epsilon - greedy$ are used. The application of this strategy results in the random exploration of the state and action spaces to gather knowledge about the task. Only when enough information is discovered from the environment does the algorithm's behavior improve. Such random exploration policies, however, waste a significant amount of time exploring irrelevant regions of the state and action spaces in which the optimal policy will never be encountered. This problem is compounded in domains with extremely large and continuous state and action spaces in which random exploration will never likely visit the regions of the spaces necessary to learn (near-)optimal policies. Additionally, in many real RL tasks with real robots, a random exploration to gather information from the environment cannot even be applied. With real robots, what is considered to be sufficient information can be much more information than a real robot can gather from the environment. Finally, as it is impossible to avoid undesirable situations in high-risk environments without a certain amount of prior knowledge about the task, the use of random exploration would require that an undesirable state be visited before it can be labeled as undesirable. However, such visits to undesirable states may result in damage or injury to the agent, the learning system or external entities. Consequently, visits to these states should be avoided from the earliest steps of the learning process.

Mitigating the difficulties described above, finite sets of teacher-provided examples or demonstrations can be used to incorporate prior knowledge into the learning algorithm. This teacher knowledge can be used in two general ways, either (i) to bootstrap the learning algorithm (i.e., a sort of initialization procedure) or (ii) to derive a policy from such examples. In the first case, the learning algorithm is provided with examples or demonstrations from which to bootstrap the value function approximation and lead the agent through the more relevant regions of the space. The second way in which teacher knowledge can be used refers to Learning from Demonstration (LfD) approaches in which a policy is derived from a finite set of demonstrations provided by a teacher. The principal drawback to this approach, however, is that the performance of the derived policy is heavily limited by teacher ability. While one way to circumvent the difficulty and improve performance is by exploring beyond what is provided in the teacher demonstrations, this again raises the question of how the agent should act when it encounters a state for which no demonstration exists (an unknown state).

### 2.3.2 Supporting the Exploration Process

While furnishing the agent with initial knowledge helps mitigate the problems associated with random exploration, this alone is not sufficient to prevent the undesirable situations that arise in the subsequent explorations undertaken to improve learner ability. An additional mechanism is necessary to guide this subsequent exploration process in such a way that the agent may be kept far away from catastrophic states. In this paper, a teacher, rather than the policy derived from the current value function approximation is used for the selection of actions in unknown states. One way to prevent the agent from encountering unknown states during the exploration process would be by requesting from the beginning a teacher demonstration for every state in the state space. However, such a strategy is not possible due to (i) its computational infeasibility given the extremely large number of states in the state space and (ii) the fact that the teacher should not be forced to give an action for every state, given that many states will be ineffectual for learning the optimal policy. Consequently, PI-SRL requests teacher action only when such action is actually required (i.e., when the agent is in an unknown state).

As this paper supposes that such a teacher is available for the task to be learned, the teacher is taken as the baseline behavior. Although some studies have examined the use of robotic teachers, hand-written control policies and simulated planners, the great majority to date have made use of human teachers. This paper uses suboptimal automatic controllers as teachers, with $\pi_T$ taken as the teacher's policy.

**Definition 7** *(**Baseline behavior**). Policy $\pi_T$ is considered the baseline behavior about which three assumptions are made: (i) it is able to provide safe demonstrations of the task to be learnt from which prior knowledge can be extracted; (ii) it is able to support the subsequent exploration process, advising suboptimal actions in unknown states to reduce the probability of entering into error states and return the system to a known situation; and (iii) its performance is far from optimal.*

While optimal baseline behaviors are certainly ideal to behave safely, non-optimal behaviors are often easy (or easier) to implement or generate than optimal ones. The PI-SRL algorithm uses the baseline behavior $\pi_T$ in two different ways. First, it uses the safe demonstrations of $\pi_T$ to provide prior knowledge about the task. In this step, the algorithm builds the initial known space of the agent derived from the safe case-based policy $\pi_B^\theta$ with the purpose of mimicking $\pi_T$ through $\pi_B^\theta$. In the second step, PI-SRL uses $\pi_T$ to support the subsequent exploration process conducted to improve the abilities of the previously-learnt $\pi_B^\theta$. As the exploration process continues, an action of $\pi_T$ is requested only when required, that is, when the agent is in an unknown state (Figure 4). In this step, $\pi_T$ acts as a backup policy in the case of an unknown state with the intention of guiding the learning away from catastrophic errors or, at least, reducing their frequency. It is important to note that the baseline behavior cannot demonstrate the correct action for every possible state. However, while the baseline behavior might not be able to indicate the best action in all cases, the action it supplies should, at the very least, be safer than that obtained through random exploration.

## 2.4 The Risk Parameter

In order to maximize exploration safety, it seems advisable that movement through the state space not be arbitrary, but rather that known space be expanded only gradually by starting from a known state. Such an exploration is carried out through the perturbation of the state-action trajectories generated by the policy $\pi_B^\theta$. Perturbation of the trajectories is accomplished by the addition of Gaussian random noise to the actions in $B$ in order to obtain new ways of completing the task. Thus, the Gaussian exploration takes place around the current approximation of the action $a_i$ for the current known state $s_c \in \Omega$, with $c_i = (s_i, a_i, V(s_i))$ and $d(s_c, s_i) = \min_{1 \le j \le \eta} d(s, s_j)$. The action performed is sampled from a Gaussian distribution with the mean at the action output given by the instance selected in $B$. When $a_i$ denotes the algorithm action output, the probability of selecting action $a_i'$, $\pi(s, a_i')$ is computed using Equation 4.

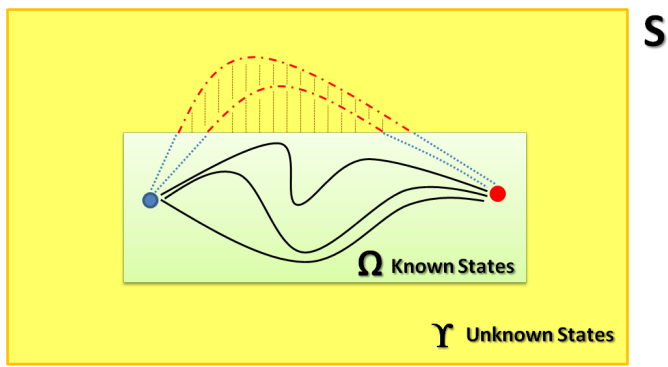

Figure 4: The exploration process in PI-SRL requests actions of the baseline behavior, $\pi_T$, when it is really required.

$$\pi(s, a_i') = \frac{1}{\sqrt{2\pi\sigma^2}} e^{-(a_i' - a_i)^2/2\sigma^2} \quad \text{if } \sigma^2 > 0. \tag{4}$$

The shape of the Gaussian distribution depends on parameter $\sigma$ (standard deviation). In this study, $\sigma$ is used as a *width parameter*. While large $\sigma$ values imply a wide bell-shaped distribution, increasing the probability of selecting actions $a_i'$ very different from the current action $a_i$, a small $\sigma$ value implies a narrow bell-shaped distribution, increasing the probability of selecting actions $a_i'$ very similar to the current action $a_i$. When $\sigma^2 = 0$, we assume $\pi(s, a_i) = 1$. Hence, the $\sigma$ value is directly related to the amount of perturbation added to the state-action trajectories generated by the policy $\pi_B^\theta$. Higher $\sigma$ values imply greater perturbations (more Gaussian noise) and a greater probability of visiting unknown states.

**Definition 8 (Risk Parameter).** *The parameter $\sigma$ is considered a risk parameter. Large values of $\sigma$ increase the probability of visiting distant unknown states and, hence, increase the probability of reaching error states.*

These exploratory actions drive the agent to the edge of the known space and force it to go slightly beyond, into the unknown space, in search of better, safer behaviors. After a period of time, the execution of these exploratory actions increases the known space and improves the abilities of the previously-learned safe case-based policy $\pi_B^\theta$. The risk parameter $\sigma$, as well as $\theta$, are design parameters that must be selected by the user. In Section 3.3, guidelines for this selection are offered.

It is important to note that the approach proposed in this study is based on two logical assumptions in RL derived from the following generalization principles (Kaelbling, Littman, & Moore, 1996; Sutton & Barto, 1998):

(i) **Nearby states have similar optimal actions**. In continuous state spaces, it is impossible for the agent to visit every state and store its value (or optimal action) in a table. This is why generalization techniques are needed. In large, smooth state spaces, similar states are expected to have similar values and similar optimal actions. Therefore, it is possible to use experience gathered from the environment with a limited subset of the state space and produce a reliable approximation over a much larger subset (Boyan, Moore, & Sutton, 1995; Hu, Kostiadis, Hunter, & Kalyviotis, 2001; Fernández & Borrajo, 2008). One must also note that, in the proposed domains, an optimal action is also considered to be a safe action in the sense that it never produces error states (i.e., no action is considered optimal that leads the agent to a catastrophic situation).

(ii) **Similar actions in similar states tend to produce similar effects**. Considering a deterministic domain, the action $a_t$ performed in state $s_t$ always produces the same state $s_{t+1}$. In a stochastic domain, it is understood intuitively that the execution of the action $a_t$ in state $s_t$ will produce similar effects (i.e., it produces states $\{s_{t+1}^1, s_{t+1}^2, s_{t+1}^3, \ldots\}$ where $\forall i, j \ i \neq j \ dist(s_{t+1}^i, s_{t+1}^j) \approx 0$). Additionally, the execution of the action $a_t' \sim a_t$ in a state $s_t' \sim s_t$ produces states $\{s'^1_{t+1}, s'^2_{t+1}, s'^3_{t+1}, \ldots\}$ where $\forall i, j \ dist(s'^i_{t+1}, s_{t+1}^j) \approx 0$. As explained earlier, the present study uses Euclidean distance as a similarity metric, as it has been proven successful in the proposed domains. As a result of this assumption, approximation techniques can be used, such that actions that generate similar effects can be grouped together as one action (Jiang, 2004). In continuous action spaces, the need for generalization techniques is even greater (Kaelbling et al., 1996). In this paper, the assumption also allows us to assume that low values of $\sigma$ increase the probability of visiting known states and, hence, of exploring less and taking less risks, while greater values of $\sigma$ increase the probability of reaching error states.

## 3. The PI-SRL Algorithm

The PI-SRL algorithm is composed of two main steps described in detail below.

### 3.1 First Step: Modeling Baseline Behaviors by CBR

The first step of PI-SRL is an approach for behavioral cloning, using CBR to allow a software agent to behave in a similar manner to a teacher policy (baseline behavior) $\pi_T$ (Floyd et al., 2008). Whereas LfD approaches are named differently according to what is learned (Argall et al., 2009), to prevent terminological inconsistencies here, we consider behavioral cloning (also known as imitation learning) to be an area of LfD whose goal is the reproduction/mimicking of the underlying teacher policy $\pi_T$ (Peters, Tedrake, Roy, & Morimoto, 2010; Abbott, 2008).

When using CBR for behavioral cloning, a case can be built using the agent's state received from the environment, as well as the corresponding action command performed by the teacher. In PI-SRL, the objective of the first step is to properly imitate the behavior of $\pi_T$ using the cases stored in a case-base. At this point, an important question arises; namely, how a case-base $\pi_B$ can be learnt using the sample trajectories provided by $\pi_T$ such that, at the end of the learning process, the resulting policy derived from $\pi_B$ mimics the behavior of $\pi_T$? Baseline behavior is a function that maps states to actions $\pi_T : S \rightarrow A$ or, in other words, a function that, given a state $s_i \in S$, provides the corresponding action $a_i \in A$. In this paper, we want to build a policy $\pi_B$ derived from a case-base composed of cases $(s_j, a_j)$ such that, for a new state $s_q$, the case with the minimum Euclidean distance $dist(s_q, s_j)$ is retrieved and the corresponding action $a_j$ is returned. Intuitively, it can be assumed that $\pi_B$ can be built simply by storing all cases $(s_i, a_i)$ gathered from one interaction between $\pi_T$ and the environment during a limited number of episodes $K$. At the end of $K$ episodes, one expects the resulting $\pi_B$ to be able to properly mimic the behavior of $\pi_T$. However, informal experimentation in the helicopter hovering domain shows this not to be the case (Section 4.3). In helicopter hovering, after $K = 100$ episodes and the prohibitive number of 600,000 cases stored, the policy derived from the case-base $\pi_B$ is unable to correctly imitate the baseline behavior $\pi_T$ and, instead, continuously crashes the helicopter. Indeed, in order for $\pi_B$ to mimic $\pi_T$ in large continuous and stochastic domains, the approach requires a larger number of episodes and, consequently, a prohibitive number of cases. In fact, to perfectly mimic $\pi_T$ in these domains, an infinite number of cases would be required. Figure 5 attempts to explain why we believe that this learning process does not work. In it, the region of the space represented by simply storing cases derived from $\pi_T$ in the form $c = (s, a)$ is shown. Each stored case (red circles) covers an area of the space and represents the centroid of a Voronoi region.

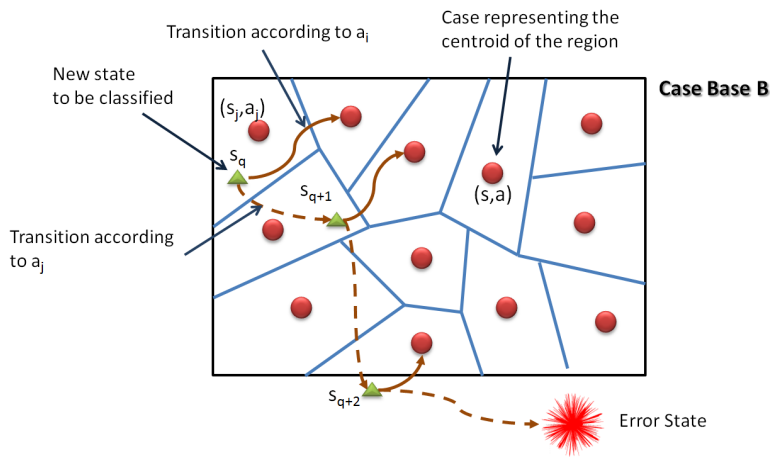

Figure 5: Effects of storing all training cases.

If the previously-learned policy $\pi_B$ is used when a new state $s_q$ is presented, the action $a_j$ is performed, corresponding to the case $c_j = (s_j, a_j)$ where the Euclidean distance $dist(s_q, s_j)$ is less than that with all other stored cases. However, if we use the policy $\pi_T$ to provide an action in the situation $s_q$, the action $a_i$ is provided which is different than $a_j$. At this point, the policy $\pi_B$ can be said to classify the state $s_q$ as the *obtained class* $a_j$, while the policy $\pi_T$ can be said to classify the state $s_q$ as the *desired class* $a_i$ (insofar as $\pi_T$ is the policy to be mimicked), with $|a_i - a_j| > 0$. Furthermore, $|a_i - a_j|$ is understood as the *classification error*. If the case-base stored all the possible pairs $(s_i, a_i)$ that $\pi_T$ were able to generate in the domain, the actions $a_j$ and $a_i$ would always be identical, with $dist(s_q, s_j) = 0$ and $|a_i - a_j| = 0$. However, in a stochastic and large, continuous domain, it is impossible to store all such cases. The sum of all such classification errors in an episode leads to the visiting of unexplored regions of the case space (i.e., regions where the new state $s_q$ received from the environment has a Euclidean distance $dist(s_q, s_j) >> \theta$ with respect to the closest case $c_j = (s_j, a_j)$ in $B$). When these unexplored regions are visited, the difference between the obtained class derived from $\pi_B$ and the desired class derived from $\pi_T$ is large (i.e., $|a_i - a_j| >> 0$) and the probability that error states might be visited greatly increases.

It may be concluded, therefore, that simply storing the pairs $c = (s, a)$ generated by $\pi_T$ is not sufficient to properly mimic its behavior. For this reason, the algorithm in Figure 6 below has been proposed.

| **CBR Approach for Behavioral Cloning** |
|---|
| 00    Given the baseline behavior $\pi_T$ |
| 01    Given the density threshold $\theta$ |
| 02    Given the maximum number of cases $\eta$ |
| 03    1. Set the case-base $B = \emptyset$ |
| 04    **2. Repeat** |
| 05      Set $k = 0$ |
| 06      **while** $k < maxEpisodeLength$ **do** |
| 07        Compute the case $< s_c, a_c, 0 >$ closest to the current state $s_k$ |
| 08        **if** $\varrho^{\pi_B^\theta}(s_k) = 0$ **then** // By equation 2 |
| 09          Set $a_k = a_c$ |
| 10        **else** |
| 11          Set $a_k$ using the **baseline behavior** $\pi_T$ |
| 12          Create a new case $c^{new} = (s_k, a_k, 0)$ |
| 13          $B := B \cup c^{new}$ |
| 14        Execute $a_k$, and receive $s_{k+1}$ |
| 15        Set $k = k + 1$ |
| 16      **end while** |
| 17      **if** $\|B\| > \eta$ **then** |
| 18        Remove the $\eta - \|B\|$ least-frequently-used cases in $B$ |
| 19      **until** stop criterion becomes true |
| 20    **3. Return $B$** performing the **safe case-based policy** $\pi_B^\theta$ |

Figure 6: CBR algorithm for behavioral cloning.

In the first step of the algorithm, the state-value function $V^{\pi_B^\theta}(s_i)$ is initialized to 0 (see line 07). The value $V^{\pi_B^\theta}(s_i)$ for each case is computed in the second step of the algorithm in Section 3.2. Additionally, this step uses the case-based risk function (Equation 2) to determine whether a new state $s_k$ should be considered risky (line 08). If the new state is not risky (i.e., it is a known state $s_k \in \Omega$), a 1-nearest neighbor strategy is followed (line 09). Otherwise, the algorithm performs the action $a_k$ using the baseline behavior $\pi_T$ and a new case $c^{new} = (s_k, a_k, 0)$ is built and added to the case-base $B$ (line 13). Starting with an empty case-base, the learning algorithm continuously increases its competence by storing new experiences. However, there are a number of reasons why the inflow of new cases should be limited. Large case-bases increase the time required to find the closest cases to a new example. While this may be partially solved using techniques to reduce the retrieval time (e.g., k-*d trees* that have been used in this work), they nevertheless do not reduce the storage requirements. Several approaches to the removal of ineffectual cases during training exist, including Aha's IBx algorithms (Aha, 1992) or any nearest prototype approach (Fernandez & Isasi, 2008). When the number of cases stored in $B$ exceeds a critical value $\|B\| > \eta$ such that the realization of a retrieval within a certain amount of time cannot be guaranteed, the removal of some cases is inevitable. An efficient approach to such a problem is through the removal of the least-frequently-used elements of $B$ (line 18).

The result of this step is a constrained case-base $B$ describing the safe case-based policy $\pi_B^\theta$ that mimics the baseline behavior $\pi_T$, though perhaps with some deviation (line 20). Formally, let $U(\pi_T)$ be an estimate of the utility of the baseline behavior $\pi_T$ computed by averaging the sum of rewards accumulated in each of $N_T$ trials. Then, $U(\pi_B^\theta) \leq U(\pi_T)$.

## 3.2 Second Step: Improving the Learned Baseline Behavior

In this step of the PI-SRL algorithm, the safe case-based policy $\pi_B^\theta$ learned in the previous step is improved by the safe exploration of the state-action space. First, for each case $c_i \in B$, the state-value function $V^{\pi_B^\theta}(s_i)$ is computed following a Monte Carlo (MC) approach (Figure 7).

| | **MC Algorithm Adapted to CBR** |
|---|---|
| 00 | Given the case-base $B$ |
| 01 | 1. Initialize, **for each** $c^i \in B$ |
| 02 | $V(s) \leftarrow arbitrary$ |
| 03 | $Returns(s) \leftarrow empty\ list$ |
| 04 | 2. **while** $k < maxNumberEpisodes$ |
| 05 | Generate an episode using $\pi_B^\theta$ |
| 06 | **for each** $s$ appearing in the episode with $< s, a, V(s) > \in B$ |
| 07 | $R \leftarrow$ return following the first occurrence of $s$ |
| 08 | Append $R$ to $Returns(s)$ |
| 09 | $V(s) \leftarrow average(Returns(s))$ |
| 10 | Set $k = k + 1$ |
| 11 | 3. **Return** $B$ |

Figure 7: Monte Carlo algorithm for the computation of state-value function for each case.

This algorithm is similar in spirit to a first-visit MC method for $V^\pi$ (Sutton & Barto, 1998), adapted in this paper to work with a policy given by a case-base. In the algorithm shown in Figure 7, all returns for each state $s_i \in B$ are accumulated and averaged, following the policy $\pi_B^\theta$ derived by the case base $B$ (see line 09). It is important to note that in the algorithm the term *return* following the first occurrence of $s$ refers to the expected return of $s$ (i.e., the expected cumulative future discounted reward starting from that state), whereas *Returns* refers to a list composed of each *return* of $s$ in different episodes. One of the principal reasons for using the MC method is that it allows us to quickly and easily estimate state values $V^{\pi_B^\theta}(s_i)$ for each case $c_i \in B$. In addition, MC methods have been shown to be successful in a wide variety of domains (Sutton & Barto, 1998). Once the state-value function $V^{\pi_B^\theta}(s_i)$ is computed for each case $c_i \in B$, small amounts of Gaussian noise are randomly added to the actions of the policy $\pi_B^\theta$ in order to obtain new and improved ways to complete the task. The algorithm used to improve the baseline behavior learned in the previous step is depicted in Figure 8. The algorithm is composed of four steps performed in each episode.

- **(a) Initialization step**. The algorithm initializes the list used to store cases occurring during an episode and sets the cumulative reward counter of the episode to 0.

- **(b) Case Generation**. The algorithm builds a case for each step of an episode. For each new state $s_k$, the closest case $< s, a, V(s) > \in B$ is computed using the Euclidean distance metric from Equation 1 (see line 09 in algorithm of Figure 8). In order to determine the perceived degree of risk of the new state $s_k$, the case-based risk function is used (line 10). If $\varrho^{\pi_B^\theta}(s_k) = 0$, then $s_k \in \Omega$ (known state). In this case, the action $a_k$ performed is computed using Equation 4 and a new case $c^{new} = < s, a_k, V(s) >$ is built to be added to the list of cases having occurred in the episode (line 13). It is important to note that the new case $< s, a_k, V(s) >$ is built replacing the action $a$ corresponding to the closest case in $< s, a, V(s) > \in B$, with the new action $a_k$ resulting from the application of random Gaussian noise to $a$ in the Equation 4. Thus, the algorithm only produces smooth changes in the cases of $B$ where $a_k \sim a$. If, however, $\varrho^{\pi_B^\theta}(s_k) = 1$, the state $s_k \in \Upsilon$ (i.e., unknown state [line 14]). In unknown states, the action $a_k$ performed is suggested by the baseline behavior $\pi_T$ which defines safe behavior (line 15). A new case $< s_k, a_k, 0 >$ is built and added to the list of cases in the episode and actions will be performed using $\pi_T$ until the agent is not in a known state. Finally, the reward obtained in the episode is accumulated, where $r(s_k, a_k)$ is the immediate reward obtained when action $a_k$ is performed in state $s_k$ (line 18).

- **(c) Computing the state-value function for the unknown states**. In this step, the state-value function of the states considered to be unknown in the previous step is computed. In the previous step (line 17), the state-value function for these states is set at 0. The algorithm proceeds in a manner similar to the first-visit MC algorithm in Figure 7. In this case, the return for each unknown state $s_i$ is computed, but not averaged since only one episode is considered (line 24 and 25). The return for each $s_i$ is computed, taking into account the first visit of the state $s_i$ in the episode (each occurrence of a state in an episode is called a visit to $s_i$), although the state $s_i$ could appear multiple times in the rest of the episode.

- **(d) Updating the cases in $B$ using experience gathered**. Updates in $B$ are made with the cases gathered from episodes with a cumulative reward similar to that of the best episode found to that point using the threshold $\Theta$ (line 27). In this way, *good sequences* are provided for the updates since it has been shown that such sequences of experiences can cause an adaptive agent to converge to a stable and useful policy, whereas *bad sequences* may cause an agent to converge to an unstable or bad policy (Wyatt, 1997). This also prevents the degradation of the initial performance of $B$ as computed in the first step of the algorithm through the use of bad episodes, or episodes with errors, for updates. In this step, two types of updates appear, namely, replacements and additions of new cases. Again, the algorithm iterates for each case $c_i = (s_i, a_i, V(s_i)) \in listCasesEpisode$ (line 29). If $s_i$ is a known state (line 30), we compute the case $< s_i, a, V(s_i) > \in B$ corresponding to the state $s_i$ (line 31). One should note that the case $c_i = (s_i, a_i, V(s_i)) \in listCasesEpisode$ was built in line 13 of the algorithm, replacing the action $a$ corresponding to the case $< s_i, a, V(s_i) > \in B$ with the new action $a_i$ and resulting from the application of random Gaussian noise to the action $a$ by the Equation 4. Then, the temporal distance (TD) error $\delta$ is computed (line 32). If $\delta > 0$, performing the action $a_i$ results in a positive change for the value of a state. The action, in turn, could potentially lead to a higher return and, thus, to a better policy. Van Hasselt and Wiering (2007) also update the value function using only the actions that potentially lead to a higher return. If the TD error $\delta$ is positive, $a_i$ is considered to be a good selection and is reinforced. In the algorithm, this reinforcement is carried out by updating the output of the case $<s_i, a, V(s_i)> \in B$ at $a_i$ (line 34). Therefore, an update to the case-base only occurs when the TD error is positive. This is similar to a linear reward-inaction update for learning automata (Narendra & Thathachar, 1974, 1989) in which the sign of the TD error is used as a measure of success. PI-SRL only updates the case-base when actual improvements have been observed, thus avoiding slow learning when there are plateaus in the value space and TD errors are small. It has been shown empirically that this procedure can result in better policies than when step size depends on the size of the TD error (Van Hasselt & Wiering, 2007). It is important to note that these replacements produce smooth changes in the case-base $B$ since an action $a$ is replaced only if $a_i$ results in a higher $V(s_i)$ and $a_i \sim a$. This form of updating can be understood as a *risk-seeking* approach, overweighting only transitions to successor states that promise an above-average return (Mihatsch & Neuneier, 2002). Additionally, it prevents the degradation of $B$, ensuring that replacements are made only when an action can potentially lead to a higher $V(s_i)$.

| | |
|---|---|
| **Policy Improvement Algorithm** | |
| 00 | Given the case-base $B$, and the maximum number of cases $\eta$ |
| 01 | Given the baseline behavior $\pi_T$ |
| 02 | Given the update threshold $\Theta$ |
| 03 | 1. Set $maxTotalRwEpisode = 0$, the maximum cumulative reward reached in an episode |
| 04 | 2. **Repeat** |
| 05 | (a) ***Initialization step***: |
| 06 | **set** $k = 0$, $listCasesEpisode \leftarrow \emptyset$, $totalRwEpisode = 0$ |
| 07 | (b) ***Case generation***: |
| 08 | **while** $k < maxEpisodeLength$ **do** |
| 09 | Compute the case $< s, a, V(s) > \in B$ closest to the current state $s_k$ |
| 10 | **if** $\varrho^{\pi_B^\theta}(s_k) = 0$ **then** // known state |
| 11 | Chose an action $a_k$ using equation 4 |
| 12 | Perform action $a_k$ |
| 13 | Create a new instance $c^{new} := (s, a_k, V(s))$ |
| 14 | **else** // unknown state |
| 15 | Chose an action $a_k$ using $\pi_T$ |
| 16 | Perform action $a_k$ |
| 17 | Create a new instance $c^{new} := (s_k, a_k, 0)$ |
| 18 | $totalRwEpisode := totalRwEpisode + r(s_k, a_k)$ |
| 19 | $listCasesEpisode := listCasesEpisode \cup c^{new}$ |
| 20 | Set $k = k + 1$ |
| 21 | (c) ***Computing the state-value function for the unknown states***: |
| 22 | **for each** instance $c_i$ in $listCasesEpisode$ |
| 23 | **if** $\varrho^{\pi_B^\theta}(s_i) = 1$ **then** // unknown state |
| 24 | $return(s_i) := \sum_{j=n}^{k} \gamma^{j-n} r(s_j, a_j)$ // n is the first ocurrence of $s_i$ in the episode |
| 25 | $V(s_i) := return(s_i)$ |
| 26 | (d) ***Updating the cases in $B$ using the experience gathered***: |
| 27 | **if** $totalRwEpisode > (maxTotalRwEpisode - \Theta)$ **then** |
| 28 | $maxTotalRwEpisode := \boldsymbol{max}(maxTotalRwEpisode, totalRwEpisode)$ |
| 29 | **for each** case $c_i = < s_i, a_i, V(s_i) >$ in $listCasesEpisode$ |
| 30 | **if** $\varrho^{\pi_B^\theta}(s_i) = 0$ **then** // known state |
| 31 | Compute the case $< s, a, V(s) > \in B$ corresponding to the state $s_i$ |
| 32 | Compute $\delta = r(s_i, a_i) + \gamma V(s_{i+1}) - V(s_i)$ |
| 33 | **If** $\delta > 0$ **then** |
| 34 | Replace the case $< s, a, V(s) > \in B$ with the case $< s_i, a_i, V(s_i) > \in listCasesEpisode$ |
| 35 | $V(s_i) = V(s_i) + \alpha\delta$ |
| 36 | **else** // unknown state |
| 37 | $B := B \cup c_i$ |
| 38 | **if** $\|B\| > \eta$ **then** |
| 39 | Remove the $\eta - \|B\|$ least-frequently-used cases in $B$ |
| 40 | **until** stop criterion becomes true |
| 41 | 3. **Return** $B$ |

Figure 8: Description of step two of PI-SRL algorithm.

If, instead, $s_i$ is not a known state, the case $c_i$ is added to $B$ (line 37). Finally, the algorithm removes cases from $B$ if necessary (line 39). Complex scoring metrics to calculate which cases are to be removed for a given moment have been proposed by several authors. Forbes and Andres (2002) suggest the removal of cases that contribute least to the overall approximation, while Driessens and Ramon (2003) pursue a more error-oriented view and propose the deletion of cases that contribute most to the prediction error of other examples. The principal drawback of these more sophisticated measures is their complexity. The determination of the case(s) to be removed involves the computation of a score value for each $c_i \in B$, which in turn requires at least one retrieval and regression, respectively, for each $c_j \in B$ $(j \neq i)$. Such entire repeated sweeps through the case-base entail an enormous computational load. Gabel and Riedmiller (2005) compute a different score metric for each $c_i \in B$, requiring the computation of the set of the $k$-nearest neighbors around $c_i$. Such approaches are not well-suited to systems learning with adjusted time requirements and with a high-dimensional state space, requiring the use of larger case-bases than those proposed here. Rather, in this paper, we propose the removal of the least-frequently-used cases. The idea seems intuitive insofar as the least-frequently-used cases usually contain worse estimates of a corresponding state's value; although the strategy might lead to a function approximator that "forgets" some of the valuable experience made in the past (e.g., *corner cases*). Despite this, PI-SRL performs successfully in all domains proposed using the strategy, as demonstrated in Section 4. Thus, the ability to forget ineffectual known states described in Section 2 is a result of the algorithm removing $\|B\| - \eta$ cases from the least-frequently-used cases of $B$.

## 3.3 Parameter Setting Design

One of the main difficulties of applying the PI-SRL algorithm to a given problem is to decide on an appropriate set of parameter values for the threshold $\theta$, the risk parameter $\sigma$, the update threshold $\Theta$ and the maximum number of cases $\eta$. An incorrect value for the parameter $\theta$ can lead to mislabeling a state as known when it is really unknown, potentially leading to damage or injury in the agent. In the case of the risk parameter $\sigma$, high values can continuously result in damage or injury; while low values are safe, but do not allow for exploration of the state-action space sufficient for reaching a near-optimal policy. Unlike $\theta$ and $\sigma$, the parameter $\Theta$ is not related to risk, but instead is directly related to the performance of the algorithm. Parameter $\Theta$ is used to determine how good an episode must be with respect to the best episode obtained, since only the best episodes are used to update the case-base $B$. If the $\Theta$ value is too large, bad episodes may be used to update $B$ (influencing the convergence and performance of the algorithm). If, instead, $\Theta$ is too low, the number of updates in $B$ may be insufficient for improving the baseline behavior. Finally, a very high $\eta$ value allows for large case-bases, increasing the computational effort during retrieval and degrading the efficiency of the system. By contrast, a very low $\eta$ value might excessively restrict the size of the case-base and thus negatively affect the final performance of the algorithm. In this subsection, a solid perspective is given on the automatic definition of these parameters. The parameter setting proposed here are taken as a suitable set of heuristics tested successfully in a wide variety of domains (Section 4).

- **Parameter $\theta$**: The parameter is domain-dependent and related to the *average size* of the actions. In this paper, the value for this parameter has been established by computing the mean distance between states during an execution of the baseline behavior $\pi_T$. Expressed in another way, the execution of the policy $\pi_T$ provides a state-action sequence of the form $s_1 \rightarrow a_1 \rightarrow s_2 \rightarrow a_2 \rightarrow \ldots \rightarrow s_n$. Thus, the value of $\theta$ is computed using Equation 5.

$$\theta = \frac{dist(s_1, s_2) + \ldots + dist(s_{n-1}, s_n)}{n - 1} \tag{5}$$

- **Parameter $\sigma$**: Several authors agree that it is impossible to completely avoid all accidents (Moldovan & Abbeel, 2012; Geibel & Wysotzki, 2005). It is important to note that PI-SRL is completely safe only if the first step of the algorithm is executed. However, by proceeding in this way, the performance of the algorithm is heavily limited by the abilities of the baseline behavior. The running of the subsequent exploratory process is inevitable if learner performance is to be improved beyond that of the baseline behavior. Since the agent operates in a state of incomplete knowledge of the domain and its dynamic, it is inevitable during the exploratory process that unknown regions of the state space will be visited where the agent may reach an error state. However, it is possible to adjust the risk parameter $\sigma$ to determine the level of risk assumed during this exploratory process. In this paper, we start with low $\sigma$ values (low risk) which we gradually increase. Specifically, we propose beginning with $\sigma = 9 \times 10^{-7}$ and increasing this value iteratively until either an accurate policy is obtained or the amount of damage or injury is high.

- **Parameter $\Theta$**: The value of this parameter is set relative to the best episode obtained. In this paper, the $\Theta$ value is set to 5% of the cumulative reward of the best episode obtained.

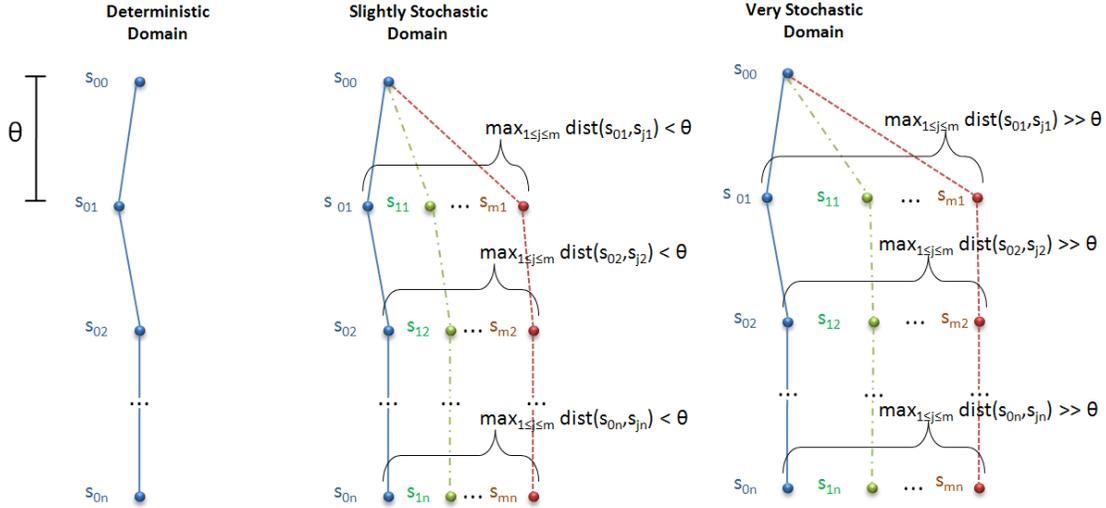

Figure 9: Trajectories generated by the baseline policy $\pi_T$ in a deterministic, slightly stochastic and highly stochastic domain.

- **Parameter** $\eta$: Previously, we estimated the maximum number of cases $\eta$ to be stored in the case-base as being the estimated maximum number of cases required to properly mimic the baseline behavior $\pi_T$. What follows is a description of how this value is computed. Figure 9 presents the trajectories (sequences of states) followed by the baseline policy $\pi_T$ in three different domains: deterministic, slightly stochastic and highly stochastic. For each domain, different sequences of the states produced by $\pi_T$ are represented $\{s_{00}, s_{01}, s_{02}, \ldots, s_{0n}\}$, $\{s_{00}, s_{11}, s_{12}, \ldots, s_{1n}\}, \ldots, \{s_{00}, s_{m1}, s_{m2}, \ldots, s_{mn}\}$, where $s_{ji}$ is the $i$-th state, $s_{00}$ the initial state and $s_{jn}$ the final state of the resulting trajectory in episode $j$. In the deterministic domain, the $m$ different executions of $\pi_T$ always result in the same trajectory. In this case, we set the maximum number of cases to $\eta = n$ with all the cases computed in the episode being stored.

In the slightly stochastic domain, the trajectories produced in $m$ different episodes are different, but only slightly so. Here, we suppose the case-base at the beginning to be empty. Additionally, we assume that all states $\{s_{00}, s_{01}, s_{02}, \ldots, s_{0n}\}$ corresponding to the first trajectory produced in the domain will be stored in the case-base. Furthermore, for each domain we execute $m$ different episodes, obtaining $m$ different trajectories. Following the execution of the $m$ episodes, we compute the maximum distance between the $i$-th state of the first trajectory (previously added to the case-base) and the $i$-th state produced in the trajectory $j$ such that $\max_{1 \le j \le m} d(s_{0i}, s_{ji})$. In the slightly stochastic domain, this maximum distance does not exceed the threshold $\theta$ in any case such that $\max_{1 \le j \le m} d(s_{0i}, s_{ji}) < \theta$. At this point, we assume the $i$-th state in trajectory $j$ to have at least one neighbor with a distance less than $\theta$ (corresponding to the state $s_{0i}$). Thus, the $i$-th state in $j$ is not added to the case-base.

By contrast, in a highly stochastic domain, this maximum distance greatly exceeds the threshold $\theta$ in all the cases such that $\max_{1 \le j \le m} d(s_{0i}, s_{ji}) >> \theta$. In this domain, we estimate the total number of cases that will be added to the case-base in the following way. For each $i$-th state in the sequence of the first trajectory, we estimate the number of cases to be added to the case-base as $\left\lfloor \frac{\max_{1 \leq j \leq m} d(s_{0i}, s_{ji})}{\theta} \right\rfloor$ or, in other words, we compute the number of intervals in the range $[0, \max_{1 \leq j \leq m} d(s_{0i}, s_{ji})]$ with a width of $\theta$ (the threshold used to decide whether a new case is to be added or not to the case-base). Consequently, the estimated number of cases added to the case-base, taking into account all states in the sequence, is computed as $\sum_{i=0}^{n} \left\lfloor \frac{\max_{1 \leq j \leq m} d(s_{0i}, s_{ji})}{\theta} \right\rfloor$. Finally, the estimated maximum number of cases is computed as shown in Equation 6.

$$\eta = n + \left( \sum_{i=0}^{n} \left\lfloor \frac{\max_{1 \leq j \leq m} d(s_{0i}, s_{ji})}{\theta} \right\rfloor \right) \qquad (6)$$

It is important to remember that in a deterministic domain, the summation in equation 6 is equal to 0 and that, therefore, $\eta = n$. The increase of the value of this element is related to the increase of stochasticity of the environment, insofar as the greater stochasticity of the environment increases the number of cases required. Finally, if the number of cases is very large or nearly infinite, the threshold $\theta$ can be increased to make more restrictive the addition of new cases to the case-base. However, this increase may also adversely affect the final performance of the algorithm.

## 4. Experimental Results

This section presents the experimental results collected from the use of PI-SRL for policy learning in four different domains presented in order of increasing complexity (i.e., increasing number of variables describing states and actions): the car parking problem (Lee & Lee, 2008), pole-balancing (Sutton & Barto, 1998), helicopter hovering (Ng et al., 2003) and the business simulator SIMBA (Borrajo et al., 2010). For each of these domains, we have proposed the learning of a near-optimal policy which minimizes car accidents, pole disequilibrium, helicopter crashes and company bankruptcies, respectively, during the learning phase. All four of the domains are stochastic in our experimentation. While both helicopter hovering and the business simulator SIMBA are, in themselves, stochastic and, additionally, generalized domains, we have made the car parking and pole-balancing domains stochastic with the intentional addition of random Gaussian noise to the actions and reward function. The results of PI-SRL in the four domains are compared to those yielded by two additional techniques, namely, the evolutionary RL approach selected winner of the helicopter domain in the 2009 RL Competition (Martín H. & Lope, 2009) and Geibel and Wysotzki's risk-sensitive RL approach (Geibel & Wysotzki, 2005). In the evolutionary approach, several neural networks cloning error-free teacher policies are added to the initial population (guaranteeing rapid convergence of the algorithm to a near-optimal policy and, indirectly, minimizing agent damage or injury). Indeed, as the winner of the helicopter domain is the agent with the highest cumulative reward, the winner must also indirectly minimize helicopter crashes insofar as these incur large catastrophic negative rewards. On the other hand, the risk-sensitive approach defines risk as the probability $\rho^{\pi}(s)$ of reaching a terminal error state (e.g., a helicopter crash ending agent control), starting at some initial state $s$. In this case, a new value function with the weighted sum of the risk probability, $\rho^{\pi}$, and value function, $V^{\pi}$, is used (Equation 7).

$$V_{\xi}^{\pi}(s) = \xi V^{\pi}(s) - \rho^{\pi}(s) \tag{7}$$

The parameter $\xi \geq 0$ determines the influence of the $V^{\pi}(s)$-values compared to the $\rho^{\pi}(s)$-values. For $\xi = 0$, $V_{\xi}^{\pi}$ corresponds to the computation of minimum risk policies. For large $\xi$ values, the original value function multiplied by $\xi$ dominates the weighted criterion. While Geibel and Wysotzki (2005) consider only finite (discretized) action sets in their study, their algorithm has been adapted here for continuous action sets. We use CBR for value and risk function approximation and a Gaussian exploration around the current action. In the experiments, for each domain, three different $\xi$ values are used, modifying the influence of the $V$-values compared to the $\rho$-values. In all cases, the goal is to improve the control policy while, at the same time, minimizing the number of episodes with agent damage or injury. In each domain, we establish different risk levels by modifying risk parameter $\sigma$ values according to the procedure described in subsection 3.3. It is important to note that one baseline behavior used to initialize the evolutionary RL approach is exactly the same as that used subsequently in the first and second step of PI-SRL. Furthermore, the case-base in the risk-sensitive approach does not begin from scratch since it is initialized with the safe case-based policy $\pi_B^{\theta}$. This makes the comparison of performances as fair as possible, but taking into account that the different techniques make its own use of the baseline behaviors.

## 4.1 Car Parking Problem

The car parking problem is represented in Figure 10 and originates from the RL literature (Cichosz, 1996). A car, represented as the rectangle in Figure 10, is initially located inside a bounded area, represented by the dark solid lines, referred to as the driving area. The goal for the learning agent is to navigate the car from its initial position into the garage, such that the car is entirely inside, in a minimum number of steps. The car cannot move outside of the driving area. Figure 10 (b) shows the two possible paths the car can take from the starting point to the garage with an obstacle in between in order to correctly perform the task. We consider the optimal policy for the car to be that which reaches the goal state in the shortest time and which, at the same time, is free of failures.

The state space of the domain is described by three continuous variables, namely, the coordinates of the center of the car $x_t$ and $y_t$ and the angle $\theta_t$ between the car's axis and the $X$ of the coordinate system. While the car can be modeled essentially with two control inputs, speed $v$ and steering angle $\phi$, let us suppose here that the car is controlled only by the steering angle (i.e., it moves at a constant speed). Thus, the action space is described by one continuous variable $a_t \in [-1, 1]$ corresponding to the turn radius, as used in the equations below. The agent receives a positive reward value of $r = (1 - \varsigma(dist(P_t, P_g))) \times 10$, where $P_t = (x_t, y_t)$ is the center of the car, $P_g = (x_g, y_g)$ is the center of the garage (i.e., the goal position) and $\varsigma$ is a normalizing function scaling the Euclidean distance $dist(P_t, P_g)$ between $P_t$ and $P_g$ to a range $[0, 1]$ when the car is inside the garage (i.e., the reward value is greater if the car is parked correctly in the center of the garage). The agent receives a reward of -1 whenever it hits the wall or obstacle. All other steps receive a reward of -0.1. Thus, the difficulty of the problem lies not only in the reinforcement delay, but also in the fact that punishments are much more frequent than positive rewards (i.e., it is much easier to hit a wall than park the car correctly). The motion of the car is described by the following equations (Lee & Lee, 2008)

$$\theta_{t+1} = \theta_t + v\tau/(l/2)\tan(\phi \times a_t), \tag{8}$$

$$x_{t+1} = x_t + v\tau\cos(\theta_{t+1}), \tag{9}$$

$$y_{t+1} = y_t + v\tau\sin(\theta_{t+1}), \tag{10}$$

where $v$ is the linear velocity of the car (assumed to be a constant value), $\phi$ is the maximum steering angle (i.e., the car can change its position by a maximum angle of $\phi$ in both directions) and $\tau$ is the simulation time step. Gaussian noise was added to the actions and rewards with a standard deviation of 0.1, since noisy interactions are inevitable in most real-world applications. Adding this noise to the actuators and the environment, we transform the deterministic domain into a stochastic domain. It is important to note that the noise added to transform the domain into a stochastic domain is independent of the Gaussian noise with standard deviation $\sigma$ (risk parameter) used to explore the state and action space in the second step of the PI-SRL algorithm. In this case, the Gaussian noise with standard deviation $\sigma$ used for exploration will be added to the noise previously added to the actuators. In this paper, $l = 4$ $(m)$, $v = 1.0$ $(m/s)$, $\phi = 0.78$ $(rad)$ and $\tau = 0.5$ $(s)$ (the driving area and obstacle dimensions are detailed in Figure 10 [a]). The initial position of the car is fixed at $x_s = 4.0$, $y_s = 4.0$ and $\theta_s = 0.26$ $(rad)$, while the goal position is $x_g = 22.5$ and $y_g = 13.5$. For this domain, we have designed a baseline behavior $\pi_T$ with an average cumulative reward per trial of 4.75.

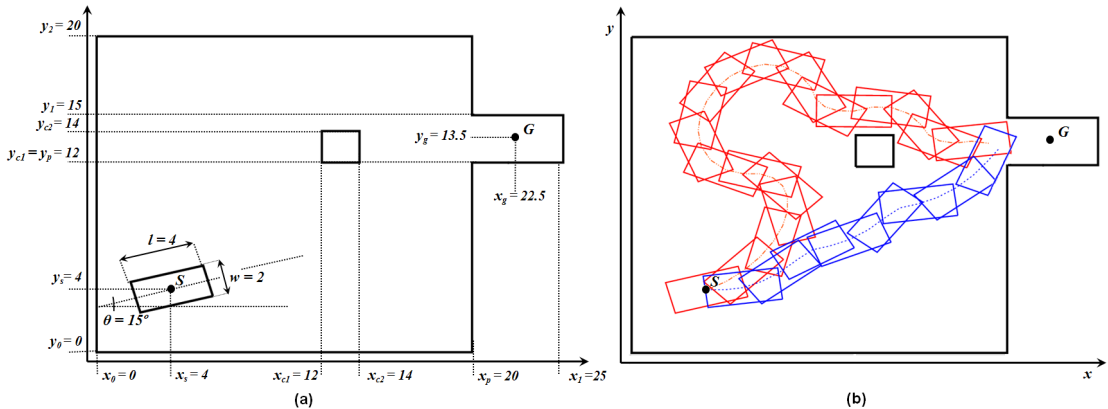

Figure 10: Car Parking Problem: (a) Model of the car parking problem. (b) Examples of trajectories generated by the agent to park the car in the garage.

In order to perform the PI-SRL algorithm, the *modeling baseline behavior step* is executed. The result of this step is the safe case-based policy $\pi_B^\theta$ learned from demonstrations provided by the baseline behavior $\pi_T$ (see subsection 3.1). $\theta$ and $\eta$ were computed following the procedure described in subsection 3.3 with resulting values of 0.01 and 207, respectively.

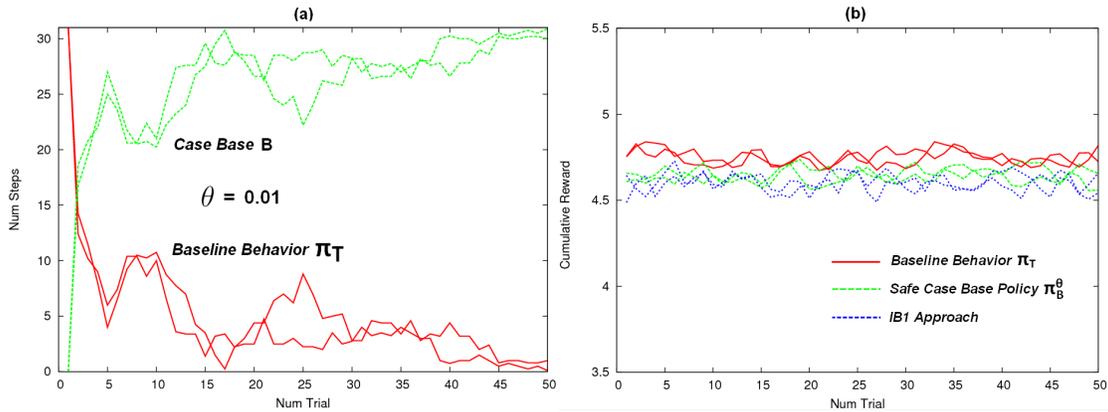

Figure 11: Car Parking Task *Modeling Baseline Behavior* Step: (a) Number of steps per trial executed by Case Base $B$ and the baseline behavior $\pi_T$. (b) Cumulative reward per trial by the baseline behavior $\pi_T$, the learned Safe Case Based Policy $\pi_B^\theta$ and an *IBL* approach.

Figure 11 (a) graphically represents the execution of the modeling baseline behavior step. In it, two different learning processes are presented and, for each one, the number of steps per trial executed by the baseline behavior $\pi_T$ (continuous red lines) and the cases in $B$ (dashed green lines) is shown. At the beginning of the learning process with an empty case-base $B$, all steps are performed using the baseline behavior $\pi_T$. As the learning process continues, new cases are added to $B$ and the safe case-based policy $\pi_B^\theta$ is learned. At around the trials 40-50, practically all steps are performed using the cases in $B$ and $\pi_T$ is rarely used, that means that a safe case-based policy has been learned. In the two learning processes shown in Figure 11 (a), the modeling baseline behavior step is performed without collisions with the wall or the obstacle. In other words, the baseline behavior $\pi_T$ is cloned safely without errors. Figure 11 (b) shows the cumulative reward for three different execution processes: the first (continuous red lines) corresponding to the performance of the baseline behavior $\pi_T$, the second (dashed green lines) corresponding to the previously-learned safe case-based policy $\pi_B^\theta$ (derived from $B$) and the third (dashed blue lines) corresponding to an instance-based learning (IBL) approach consisting of storing cases in memory. In the IBL approach, new items are classified by examining the cases stored in memory and determining the most similar case(s) given a particular similarity metric (Euclidean distance is used in this paper). The classification of that nearest neighbor (or those nearest neighbors) is taken as the classification of the new item using a 1-nearest neighbor strategy (Aha & Kibler, 1991). For each approach, two different executions are carried out. In the IBL approach, the training process is performed saving all training cases produced by the baseline behavior $\pi_T$ during 50 trials (so we consider this approach an IB*1* algorithm in the sense that it saves every case during the training phase, see Aha & Kibler, 1991). Figure 11 (b) shows that the safe case-based policy $\pi_B$ almost perfectly mimics the behavior of the baseline behavior $\pi_T$. In the domain, the performance of the IB*1* approach is also similar.

Figure 12 (a) shows the results for different risk configurations obtained by the *improving the learned baseline behavior step*. For each risk configuration, two different learning pro-cesses are performed. All trials ending in failure (car hits the wall or obstacle) are marked (blue triangles). The learning processes in Figure 12 (a) demonstrate that the number of failures increases with an increase in the parameter $\sigma$. For a low level of risk ($\sigma = 9 \times 10^{-4}$), although no failures are produced, the performance is nevertheless weak (around the baseline behavior $\pi_T$) and constant throughout the whole of the learning process. Additional experiments have demonstrated that increasing the $\sigma$ value above $\sigma = 9 \times 10^{-2}$ increases the number of failures without improving performance. Figure 12 (b) shows the results for the evolutionary and risk-sensitive RL approaches for different $\xi$ values. Regarding the former, the number of failures is higher than that obtained by the PI-SRL approach, while its final performance is similar. In the case of the latter, performance is higher when $\xi = 1.0$ (value maximization), yet the agent consistently crashes the car into the wall.

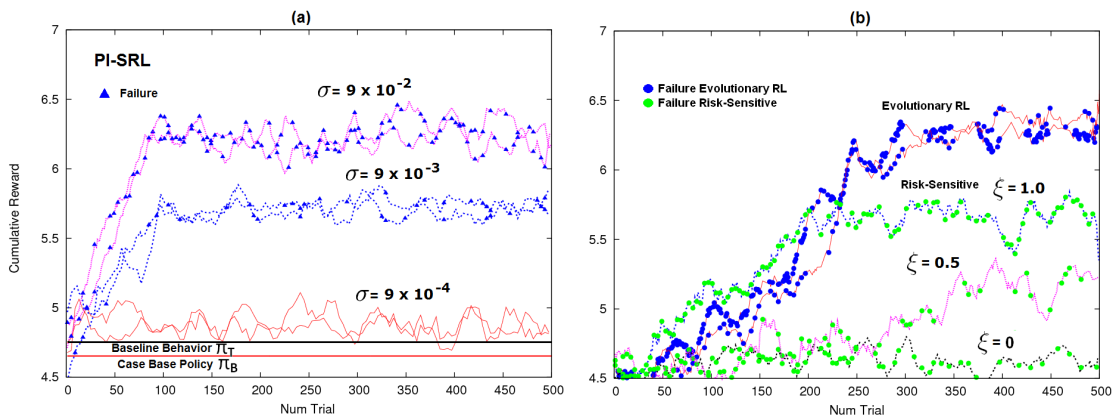

Figure 12: Improving the learned baseline behavior step in car parking problem: (a) Cumulative reward per episode for different risk configurations ($\sigma$) obtained by PI-SRL. (b) Cumulative reward per episode by the evolutionary RL and risk-sensitive RL approaches. In all cases, any episode ending in failure is marked.

Figure 13 shows the mean number of failures (i.e., car collisions) and cumulative reward for each approach over 500 trials with the red circles corresponding to the PI-SRL algorithm, the black triangles to the risk-sensitive approach and the blue square to the evolutionary RL approach. Additionally, Figure 13 shows two asymptotes. The horizontal asymptote is established according to the cumulative reward obtained by the highest $\sigma$ value. The horizontal asymptote indicates that higher $\sigma$ values increase the number of failures without improving the cumulative reward (which may, in fact, get worse). The vertical asymptote at $Failures = 0$ indicates that reducing the risk parameter $\sigma$ does not reduce the number of failures. Figure 13 also shows the performance for two additional risk levels, a very high level of risk ($\sigma = 9 \times 10^{-1}$) and very low level of risk ($\sigma = 0$), with respect to Figure 12. When using a very low level of risk $\sigma = 0$, no additional random Gaussian noise is added to the actions and the algorithm is free of failures, although performance does not improve with respect to the safe case-based policy $\pi_B^\theta$ learned in the first step of the algorithm. PI-SRL with a medium level of risk ($\sigma = 9 \times 10^{-4}$) also is free of failures, yet performance is also slightly improved. The PI-SRL algorithm with high level of risk ($\sigma = 9 \times 10^{-2}$) obtains the highest cumulative reward, 3053.37, with a mean of 78.8 failures. However, when using a very high level of risk ($\sigma = 9 \times 10^{-1}$), the number of failures greatly increases and, consequently, the cumulative reward decreases. As shown in Figure 12, PI-SRL with high risk ($\sigma = 9 \times 10^{-2}$) and the evolutionary RL approach obtain a similar performance, while PI-SRL demonstrates a faster convergence (thus, in Figure 13, the cumulative reward obtained by PI-SRL is higher). The Pareto comparison criterion can be used to compare the solutions in Figure 13. Using this principle, one solution $y^*$ strictly dominates (or "is preferred to") a solution $y$ if each parameter of $y^*$ is not strictly worse than the corresponding parameter of $y$ and at least one parameter is strictly better. This is written as $y^* \succ y$, indicating that $y^*$ strictly dominates $y$. In accordance with the Pareto principle, we can assume the points in Figure 13 corresponding to the PI-SRL solutions, save PI-SRL with very high level of risk, to be on the Pareto frontier, since these points are not strictly dominated by any other solution (i.e., no other solution has, at the same time, a higher cumulative reward and a lower number of failures than PI-SRL). In this domain, the solution of the PI-SRL with a medium level of risk strictly dominates (or "is preferred to") the risk-sensitive solutions (PI-SRL $\sigma = 9 \times 10^{-3} \succ$ risk-sensitive) and the solution PI-SRL with a high level of risk strictly dominates the solution of the evolutionary RL solution (PI-SRL $\sigma = 9 \times 10^{-2} \succ$ evolutionary RL).

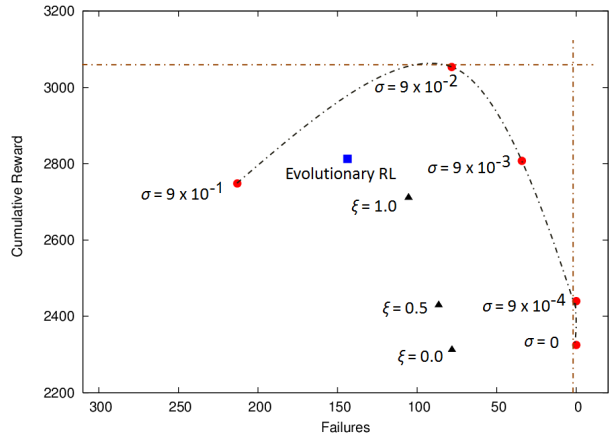

Figure 13: Mean number of failures (car collisions) and cumulative reward over 500 trials for each approach in car parking task. The means have been computed from 10 different executions.

Nevertheless, it is important to note that any ultimate decision about which approach in Figure 13 is best depends on the criteria of the researcher. If, for instance, the minimization of the number of failures is deemed the most important optimization criterion (independently of the improvement obtained with respect to the baseline behavior $\pi_T$), the best approach will be PI-SRL with a low level of risk ($\sigma = 9 \times 10^{-4}$). Similarly, if the maximization of the cumulative reward is instead judged to be the most important optimization criterion (independently of the number of failures generated), the best approach will be PI-SRL with a high level of risk ($\sigma = 9 \times 10^{-2}$).

Figure 14 shows the evolution of the cases in the case-base $B$ (known space) in different trials for a high-risk learning process. Each graph presents the set of known states $\Omega$ (green area), error states $\Phi$ (red area), unknown states $\Upsilon$ (yellow area) and non-error states $\Gamma_\Omega$ (orange circles). PI-SRL adapts the known space in order to find safer and better policies to complete the task. Figure 14 (a) shows the initial situation of $B$ (corresponding to the previously-learned safe case-based policy $\pi_B^\theta$). It is robust in the sense that it never results in any collisions, but suboptimal (it selects the longest parking path driving around the upper side of the obstacle). As the learning process progresses (Figure 14 (b)), PI-SRL finds a shorter path to park the car in the garage along the upper side of the obstacle (increasing the performance), but which comes closer to the obstacle than before (increasing the probability of collisions). In Figure 14 (c), PI-SRL finds a new and even shorter path, this time along the lower side of the obstacle. However, there are still cases in the case-base $B$ corresponding to the older path along the upper side of the obstacle (so Figure 14 (c) indicates two paths to park the car). Finally, in Figure 14 (d), the cases corresponding to the suboptimal path along the upper side of the obstacle have been removed from $B$ and replaced by new cases corresponding to the safe and improved path along the lower side of the obstacle. In other words, PI-SRL adapts the known space through the exploration of the unknown space in order to find new and improved behaviors. During this process of adjusting the known space to safe and better policies, the algorithm "forgets" the previously-learned, yet ineffective known states.

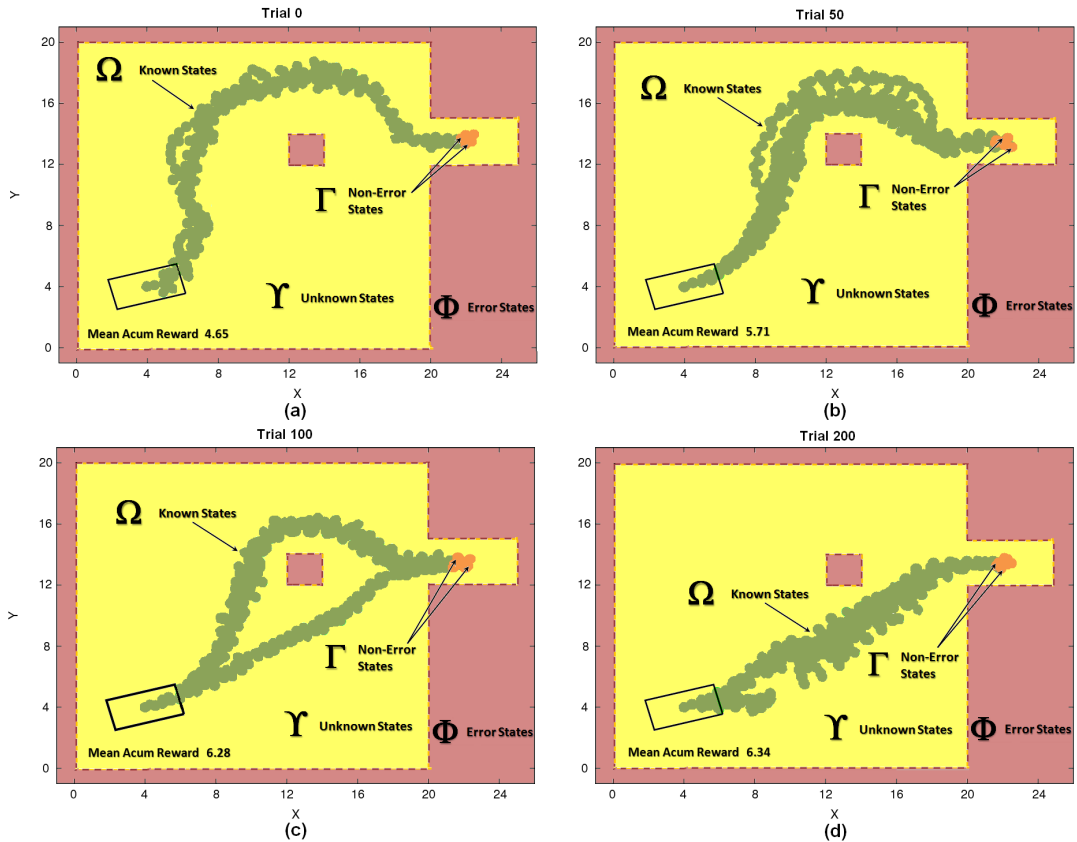

Figure 14: Car parking problem: Evolution of the known space for different trials $T = 0$ (a), $T = 50$ (b), $T = 100$ (c) and $T = 200$ (d) in a high-risk learning process ($\sigma = 9 \times 10^{-2}$). Each graph corresponds to the situation of the state space in accordance with the case-base $B$ in trial $T$.

In the following experiment, it becomes apparent that if the domain is noisy enough, even when taking no risk at all (i.e., no further noise added to the actuator for exploration), the agent could nevertheless perform poorly and constantly produce collisions. The experiment also serves to explain why domain noise can never be sufficient for the efficient exploration of the space without action selection noise. In the experiment, we have intentionally added more noise to the actuators and have performed second step of PI-SRL again, however this time taking no risk (i.e., $\sigma = 0$). In this test, we have added random Gaussian noise with a standard deviation of 0.3, rather than the standard deviation of 0.1 used previously, to the actuators. Figure 15 shows two executions of the second step (improving the learned baseline policy) of the PI-SRL algorithm with the x-axis indicating the number of trials, the y-axis the cumulative reward per episode and failures (i.e., collisions) marked as blue triangles. In the experiments in Figure 12 (b), the case-based policy $\pi_B$ with low level of risk ($\sigma = 9 \times 10^{-4}$) never produces failures. In contrast, in the experiments shown in Figure 15, the same case-based policy $\pi_B$ continually collides with the wall although the risk parameter is set to 0 ($\sigma = 0$). Furthermore, an increase in the performance can also be detected.

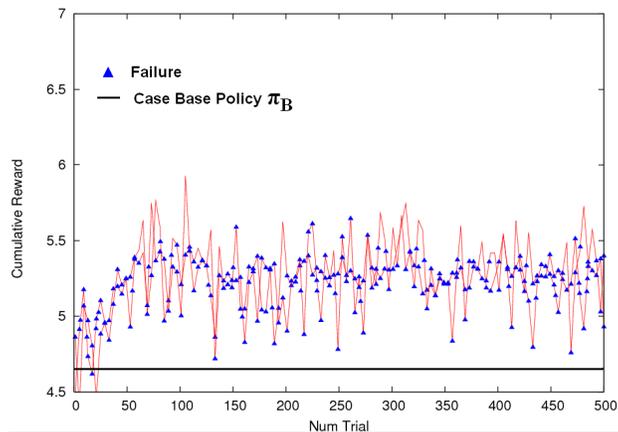

Figure 15: Improving the learned baseline behavior step of car parking task: Two learning processes for risk configuration $\sigma = 0$ and an increase in the noise in the actuators.

The increase of noise in the actuators in the second step of the algorithm with respect to the first step (the case-based policy $\pi_B$ is learned in the first step using Gaussian random noise in the actuator with a standard deviation of 0.1, while the second step is performed using Gaussian random noise in the actuator with a standard deviation of 0.3) takes the agent beyond the known space of the case-base $B$ learnt in the first step of PI-SRL and allows it to find new trajectories for parking the car in the garage. In this new situation, the exploration process is guided as follows. If a known state is reached, the agent performs the action $a$ retrieved from $B$ without the addition of Gaussian noise, since the risk parameter $\sigma = 0$ (see line 11 in Figure 8 algorithm). If an unknown state is reached, the agent performs the action $a$ advised by the baseline behavior $\pi_T$ (see line 15). Using this exploration process, if a new and better trajectory is found for parking the car in the garage, the resulting cases in the episode corresponding to unknown states are added to the case-base (see line 37), slightly improving the performance in Figure 15. It is important to note that the replacements of cases (see line 34) does not change the actions in $B$, since these are replaced by the same action previously retrieved from $B$ plus a certain amount of Gaussian noise with standard deviation $\sigma$ (see line 11). Nevertheless, given that the risk parameter $\sigma$ has been set to 0, the actions retrieved from the case-base are not replaced. This exploration process, however, with $\sigma = 0$ (i.e., taking no risk) does not lead to optimal behavior since:

- The actions performed in unknown situations and added to the case-base $B$ are performed using the baseline behavior $\pi_T$ which is supposed perform suboptimal actions (see definition of baseline behavior).

- The actions in the cases of $B$ are not replaced with improved actions. The Gaussian noise with standard deviation $\sigma$ is used to explore different and better actions than those provided by $B$ and $\pi_T$; however, in this case, the risk parameter is set to $\sigma = 0$ and new and better actions are not discovered.

Additional experiments demonstrate that PI-SRL behaves much worse when a higher value of noise is used in the actuators (with collisions in all episodes). We assume that taking no risk (i.e., $\sigma = 0$) implies always performing the same actions while not discovering any newer or better actions than those provided by the learned case-base $B$ and the baseline behavior $\pi_T$. In PI-SRL, the replacements in the case-base are executed towards the more promising action which, in our case, is that which guarantees a higher return. This is why exploration is necessary in order to obtain (near-)optimal behavior, since without exploration, new and better actions are not discovered and PI-SRL performance is limited by that of the case-based policy learned in the first step $\pi_B$ and the baseline behavior $\pi_T$ which, one must remember, is intended to perform suboptimal policies.

## 4.2 Pole-Balancing

As the name suggests, the objective in the pole-balancing problem is to balance a pole vertically on top of a moving cart (Sutton & Barto, 1998). The state description consists of a four-dimensional vector containing the angle $\phi$, the radial speed $\phi'$, the cart position $x$ and the speed $x'$. The action consists of a real-valued force that is used to push the cart. In this study, the reward is computed to encourage actions that keep the pole as upright as possible on the cart and the cart as centered as possible on the track. Thus, the reward in step $t$ is computed as $r_t = 1 - (\varsigma(\phi_t) + \rho(x_t))/2$, where $\varsigma$ and $\rho$ are normalizing functions scaling the angle $\phi_t$ and the position $x_t$ to a range $[0, 1]$. An episode is composed of 10,000 steps, although it may nevertheless end prematurely if the pole becomes unbalanced (i.e., if it has an inclination of more than twelve degrees in either direction) or the cart falls off the track (i.e., if it is more than 2.4m from the center of the track), both of which being considered failures. As in the car parking problem, Gaussian noise was added to the actions and rewards, this time with a standard deviation of $10^{-4}$. The pole-balancing domain becomes stochastic through the addition of this noise to the actuators and reward function.

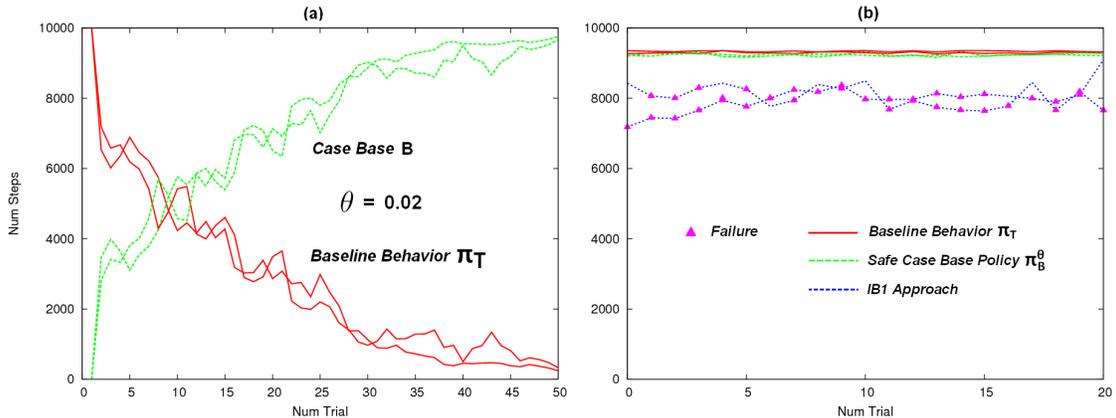

Figure 16: Modeling baseline behavior step in pole-balancing task: (a) Number of steps per trial executed by case-base $B$ and baseline behavior $\pi_T$. (b) Cumulative reward per trial for $\pi_T$, the learned safe case-based policy $\pi_B^\theta$ and an IBL approach.

The hand-made baseline behavior $\pi_T$ demonstrates the execution of a safe, yet suboptimal policy, with an average cumulative reward per episode/trial of 9292.

In the modeling baseline behavior step of PI-SRL, the safe case-based policy $\pi_B^\theta$ is learnt from demonstrations provided by the baseline behavior $\pi_T$. $\theta$ and $\eta$ were computed following the procedure described in subsection 3.3, with values of 0.02 and 12572, respectively. Figure 16 (a) shows two different learning processes for the modeling baseline behavior step. For each learning process, Figure 16 (a) shows the number of steps per trial executed by baseline behavior $\pi_T$ (continuous red lines) and by the case-base $B$ (dashed green lines). At the beginning of the learning process, the case-base $B$ is empty and all steps are performed using the baseline behavior $\pi_T$. As the learning process progresses, however, $B$ is filled and the safe case-based policy $\pi_B^\theta$ is learnt. At the end of the learning process (after around 45-50 trials), almost all steps are performed using the cases in $B$ and $\pi_T$ is rarely used. It is important to note that the modeling baseline behavior step has been performed without failures (i.e., pole disequilibrium or cart off the track) in each case. As with the previous task, Figure 16 (b) represents three independent execution processes using the previously-learned safe case-based policy $\pi_B^\theta$ (derived from $B$ and indicated with dashed green lines), the baseline behavior $\pi_T$ (indicated with continuous red lines) and an approach based on IBL (indicated with dashed blue lines) (Aha & Kibler, 1991). The average cumulative reward per episode in $\pi_B^\theta$ is 9230 (Figure 16 [b]). While $\pi_B^\theta$ almost perfectly clones $\pi_T$, the IB1 approach which, in most cases, results in pole disequilibrium or the cart falling off the track averages a cumulative reward per episode of 8055.

Figure 17 (a) shows the results of PI-SRL for different risk configurations. For each configuration, the learning curves are shown for two different learning processes performed. Additionally, any episode ending in failure is marked (blue triangles). While an increase in risk increases the probability of failure, the policy obtained is nevertheless better in terms of the cumulative reward. Nevertheless, much greater risk values ($\sigma = 9 \times 10^{-5}$) produce more failures without an accompanying increase in the cumulative reward. Figure 17 (b) shows the results for the evolutionary and risk-sensitive RL approaches, the former of which being clearly the algorithm with the greatest number of failures. In the risk-sensitive approach, for $\xi = 2.0$ (value maximization), the agent selects actions that result in a higher value, but also in a higher risk. On the contrary, for $\xi = 0$ (risk minimization), when the agent learns the risk function (at around episode 6000), it selects actions with a lower risk (and a lower number of failures), but also with considerably weak performance. The value $\xi = 0.1$ produces an intermediate policy. Consequently, it can be concluded that PI-SRL with a high level of risk obtains better policies and less failures than the evolutionary or risk-sensitive RL approaches. Figure 18 reinforces the previous conclusions.

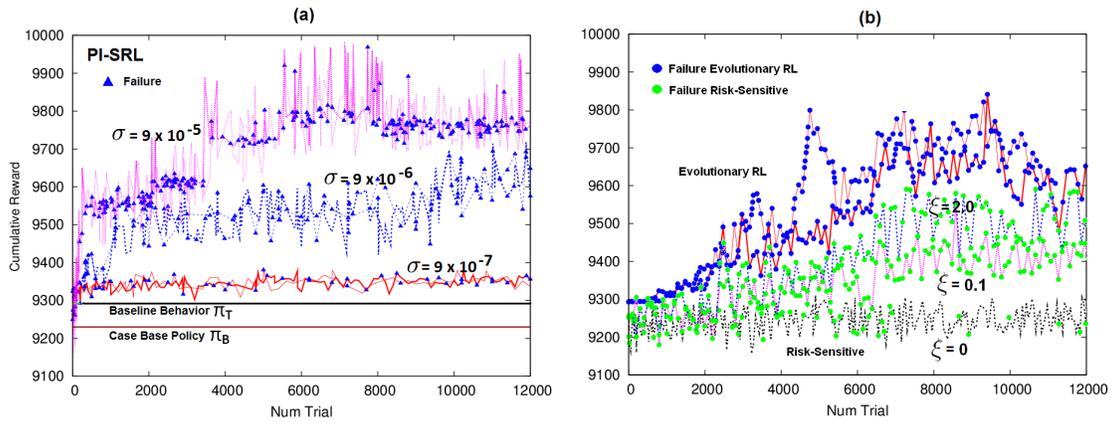

Figure 17: Improving the learned baseline behavior step of pole-balancing task: (a) Cumulative reward per episode for different risk configurations ($\sigma$) obtained by PI-SRL. (b) Cumulative reward per episode obtained by the evolutionary and risk-sensitive RL approaches. In all cases, any episode ending in failure is marked.

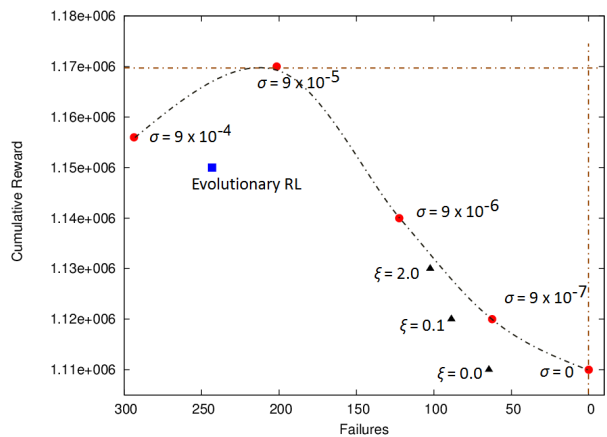

Figure 18: Mean number of failures (pole disequilibrium or cart off the track) and cumulative reward during 500 trials for each approach in the pole-balancing task. The means have been computed from 10 different executions.

In it, the mean number of failures and cumulative reward during 12,000 trials are shown, with the red circles corresponding to PI-SRL, the black triangles corresponding to the risk-sensitive approach and the blue square corresponding to the evolutionary RL approach. The figure also shows performance for two additional risk levels, a very high level of risk ($\sigma = 9 \times 10^{-4}$) and very low level of risk ($\sigma = 0$), with respect to Figure 17. The cumulative reward and number of failures increase with the high level of risk ($\sigma = 9 \times 10^{-5}$). This risk level represents an inflection point at which higher levels of risk produce more failures without an accompanying improvement in the cumulative reward. In fact, the very high level of risk ($\sigma = 9 \times 10^{-4}$) results in a reduction in the cumulative reward when compared with the high level of risk ($\sigma = 9 \times 10^{-5}$). Again, the Pareto comparison criterion may be used to compare the solutions from Figure 18. In this domain, the solution from PI-SRL with a low level of risk strictly dominates the risk-sensitive solutions with $\xi = 0.0$ and $\xi = 0.1$, such that PI-SRL $\sigma = 9 \times 10^{-7} \succ$ risk-sensitive with $\xi = 0.0$ and $\xi = 0.1$. Additionally, the solution from PI-SRL with a high level of risk strictly dominates evolutionary RL solution, such that PI-SRL $\sigma = 9 \times 10^{-5} \succ$ evolutionary RL.

Lastly, Figure 19 shows the evolution of the known space derived from the case-base $B$ in different trials for a high-risk learning process. For each graph, error states $\Phi$ (red area), the set of unknown states $\Upsilon$ (yellow area), the set of known states $\Omega$ (green area) and the set of non-error states $\Gamma_\Omega$ (orange circles) are represented. The known space $\Omega$ in each graph has been computed taking cases from $B$ in the trials $T = 0, 3000, 6000$ and $8000$. For each graph, non-error states $\Gamma_\Omega$ have been computed from 10 different executions of $B$ in the trial $T$ (the orange circles representing the terminal states for each of these executions). The first graph (Figure 19 [a]) presents the initial known space resulting from the modeling baseline behavior step. The evolution in Figure 19 demonstrates two different points. First, PI-SRL progressively adapts the known space in order to encounter better behavior such that the known space tends to be compressed toward the center of the coordinates. This is so due to the fact that the reward is greater if the angle $\phi$ of the pole and the cart position $x$ are 0 (i.e., the pole is as upright as possible on the cart and the cart is centered on the track). Second, the risk of failure in the pole-balancing domain is greater during early trials of the learning process. At the beginning of the learning process (Figure 19 [a]), $T = 0$, some regions of the known space are close to the error space. In this situation, slight modifications of the actions consistently produce visits to the states in $\Phi$ (i.e., pole disequilibrium or cart falling off the track). As the learning process advances (Figure 19 [b], [c] and [d]), the known space is compressed toward the origin of coordinates and away from the error space. Consequently, the probability of visiting error states decreases. For example, returning to Figure 17 (a), in the high-risk learning processes, 52% of the failures (126) occur in the first 4000 trials, while the remaining 48% (117) occur in the last 8000 trials.

## 4.3 Helicopter Hovering

As suggested by its name, the objective of this domain is to make a helicopter hover as close as possible to a defined position for a duration established by an episode. The task is challenging for two main reasons. Firstly, both the state and action spaces are high-dimensional and continuous (more specifically, the state space is 12-dimensional and the action space 4-dimensional). Secondly, it is a generalized domain whose behavior is modified by the wind factor. A helicopter episode is composed of 6000 steps, although it may end prematurely if the helicopter crashes. The first step of PI-SRL is performed in order to imitate the baseline behavior $\pi_T$. $\theta$ and $\eta$ were computed following the procedure described in subsection 3.3 with values of 0.3 and 49735, respectively. Once this step has been performed, the resulting safe case-based policy $\pi_B^\theta$ is able to properly imitate the baseline behavior $\pi_T$.

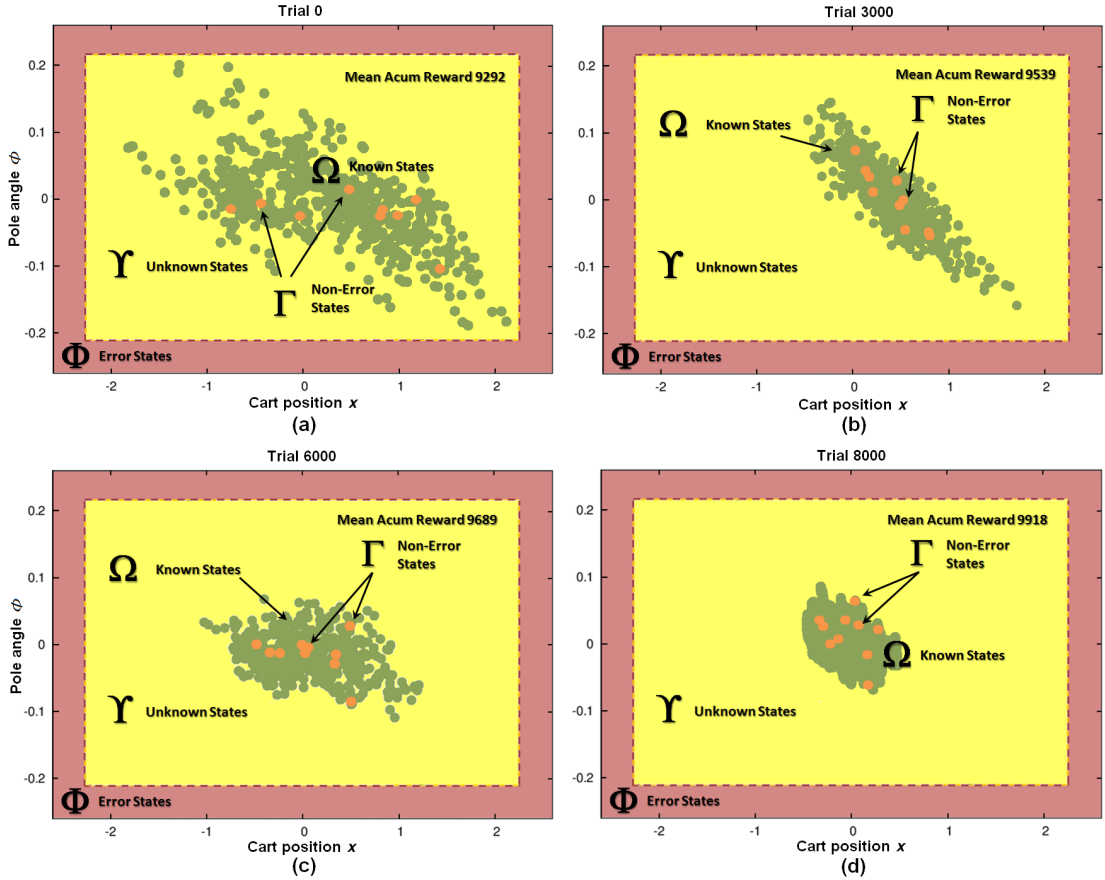

Figure 19: Pole-balancing task: Evolution of the known space for different trials $T = 0$ (a), $T = 3000$ (b), $T = 6000$ (c) and $T = 8000$ (d) in a high-risk learning process ($\sigma = 9 \times 10^{-5}$). Each graph corresponds to the situation of the state space according to the case-base $B$ in trial $T$.

Figure 20 (a) shows two learning processes of the modeling baseline behavior step. Similar to previous tasks, as the learning processes progress, the number of steps executed by the baseline behavior $\pi_T$ is reduced while the number of steps using the case-base $B$ increases. By the end of the learning process, the case-base $B$ stores the safe case-based policy $\pi_B^\theta$. Figure 20 (b) compares the performance (in terms of cumulative reward per episode) of $\pi_T$, the learned case-based policy $\pi_B^\theta$ and the IB1 approach. Regarding the first two, the average cumulative reward per episode of $\pi_T$ is -78035.93, while that obtained by $\pi_B^\theta$ is -85130.11. Although the $\pi_B^\theta$ does not perfectly mimic the baseline behavior $\pi_T$, it nevertheless performs a safe policy without crashing the helicopter. With regard to the training process of the IB$1$ approach, every case produced during 15 episodes by the baseline behavior $\pi_T$ is stored. Figure 20 (b) demonstrates that the IB$1$ approach consistently results in helicopter crashes, with a performance extremely far from that of the learned safe case-based policy $\pi_B^\theta$. Improvement of the policy $\pi_B^\theta$ begins when the state-action space is safely explored through the execution of step two of PI-SRL.

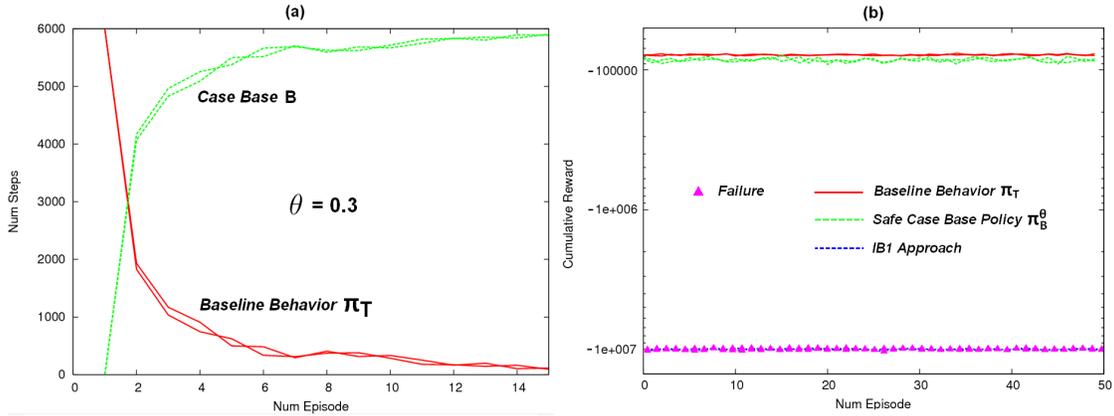

Figure 20: Modeling baseline behavior step of helicopter hovering task: (a) Number of steps per trial executed by case-base $B$ and baseline behavior $\pi_T$. (b) Cumulative reward per trial by $\pi_T$, the learned safe case-based policy $\pi_B^\theta$ and an IBL approach.

Figure 21 (a) shows the results for different risk levels. While PI-SRL low and medium levels of risk levels do not produce helicopter crashes in PI-SRL, performance is nevertheless quite weak.

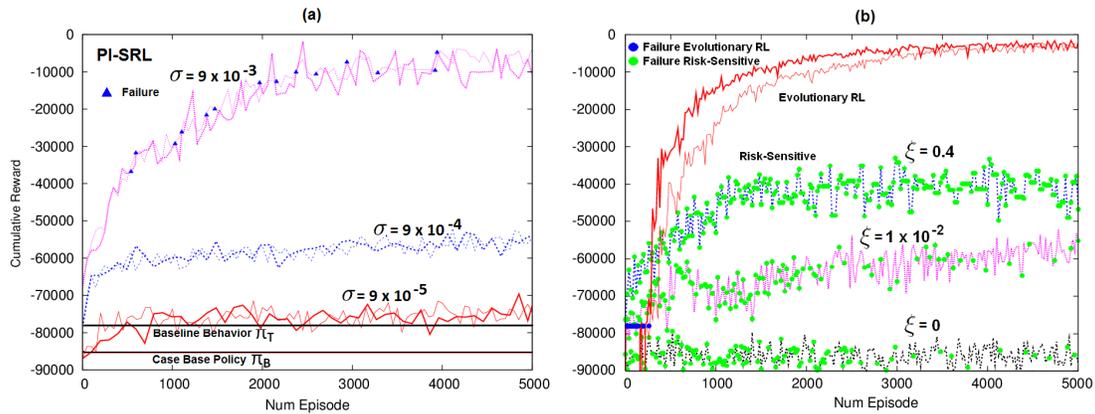

Figure 21: Improving the learned baseline behavior step in helicopter hovering task: (a) Cumulative reward per episode for different risk configurations obtained by PI-SRL. (b) Cumulative reward per episode obtained by evolutionary and risk-sensitive RL approaches. In all cases, any episode ending in failure is marked.

Conversely, the high level of risk established produces a near-optimal policy with a low number of collisions. Extensive experimentation demonstrates that increasing the risk parameter $\sigma = 9 \times 10^{-3}$ also increases the number of crashes without an accompanying improvement in the cumulative reward. Figure 21 (b) shows the results of the evolutionary RL approach which, it should be remembered, was selected winner of the RL Competition 2009 in the same domain (Martín H. & Lope, 2009), as well as the risk-sensitive RL algorithm for different $\xi$ values. A comparison of the results between the evolutionary RL approach and PI-SRL shows a similar cumulative reward, while also a significantly higher number of crashes from the former than from the latter. In the evolutionary approach, all crashes occur in the early steps of the learning process; while in PI-SRL, accidents occur at more advanced steps of the learning process. In the case of the risk-sensitive RL algorithm, for $\xi = 0$ and $\xi = 0.01$ the risk function is learned at around episode 3000. At this point, the agent selects lower-risk actions and the number of crashes is considerably reduced. When $\xi = 0.4$ and the agent selects actions resulting in higher values without taking risk into account, performance improves, but at the expense of an increased number of accidents. Nevertheless and whatever the $\xi$ value, the number of crashes is higher and the performance is worse than with PI-SRL.

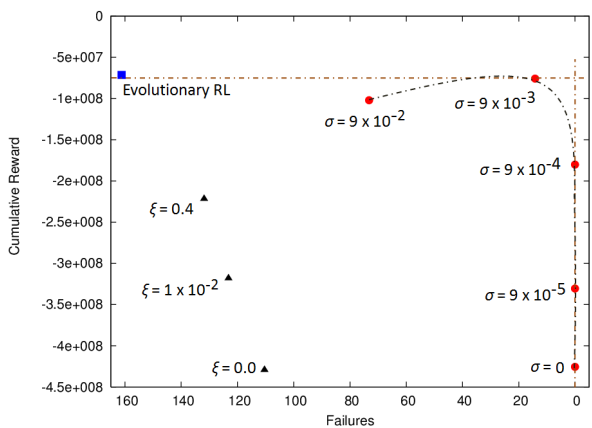

Figure 22: Mean number of failures (helicopter crashes) and cumulative reward during 5000 episodes for each approach to the helicopter hovering task. The means have been computed from 10 different executions.

The information from Figure 22, indicating the mean number of failures and cumulative reward over 5000 episodes for each approach, complements the conclusions made above. The data has been computed from 10 independent executions of each approach. As in previous domains, PI-SRL is indicated by red circles, the risk-sensitive approach by the black triangles and the evolutionary RL approach by the blue square. Figure 22 also shows the performance for two additional risk levels, a very high level of risk ($\sigma = 9 \times 10^{-2}$) and a very low level risk ($\sigma = 0$), with respect to Figure 21. Figure 22 demonstrates that the evolutionary RL approach obtains the highest cumulative reward ($-7.13 \times 10^7$), followed closely by PI-SRL ($-7.57 \times 10^7$). The other approaches are far from these results. Regarding the number of failures (i.e., helicopter crashes), as PI-SRL with a very low level of risk ($\sigma = 0$), a low level of risk ($\sigma = 9 \times 10^{-5}$) and a medium level of risk ($\sigma = 9 \times 10^{-4}$) produces no collisions, the PI-SRL algorithm with medium risk is preferable inasmuch as the cumulative reward is higher $(-18.01 \times 10^7)$. Using the Pareto comparison criterion, the PI-SRL solution with a high level of risk strictly dominates the solutions of the risk-sensitive approach (PI-SRL $\sigma = 9 \times 10^{-3} \succ$ risk-sensitive). Moreover, PI-SRL is not strictly dominated by any other solution.

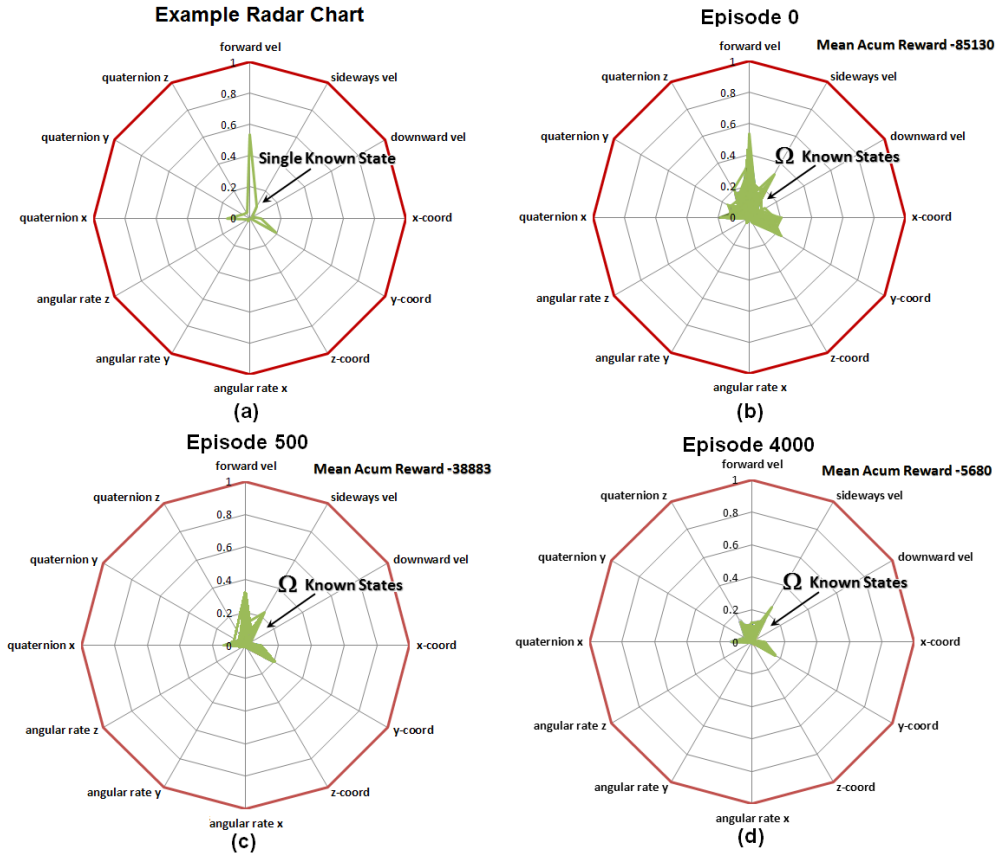

Figure 23: Evolution of the known space for different episodes in the helicopter hovering task. (a) Example of representation of a single known state in a radar chart. (b), (c), and (d) Known states in episodes $T = 0$, $T = 500$ and $T = 4000$, respectively, in a high-risk learning process ($\sigma = 9 \times 10^{-3}$). Each graph corresponds to the situation of the known space according to the case-base $B$ in episode $T$.

As with the pole-balancing domain, Figure 23 shows the evolution of the known space according to the case-base $B$ in different episodes for a high-risk learning process. In this case, radar charts are used due to the high number of features describing the states. A radar chart is a graphical method for displaying multivariate data two-dimensionally. In the Figure, each axis represents one of the features of the state and, to preserve the simplicity of the representation, the charts are generated normalizing the absolute values of the features between 0 and 1. Figure 23 (a) is an example of a representation of a single known state.

The value of each axis corresponds to the value of an individual feature in a state and a line is drawn connecting the feature values for each axis. While the line in Figure 23 (a) represents a single state, Figures 23 (b), (c) and (d) show the known space according to the case-base $B$ in episodes 0, 500 and 4000, respectively. These three charts do not represent a single state, but rather all the states in $B$ for the corresponding episode. Thus, for each graph, the set of known states is marked $\Omega$ (green area). A state is considered an error state if a single feature value for that state is greater than 1. The limits (marked by a red line in the graphs) have been computed taking into account that the helicopter crashes if (i) the velocity along any of the main axes exceeds 5 $m/s$, (ii) the position of the helicopter is off by more than 20 $m$, (iii) the angular rate around any of the main axes exceeds $2 \times 2\pi$ $rad/s$ or (iv) the orientation is more than 30 $degrees$ from the target orientation. As with previous tasks, Figure 23 indicates two different matters. First, as the learning proceeds, the known space derived from $B$ is adjusted to the space used for better and safer policies. In the helicopter domain, the agent tries to hover the helicopter as close as possible to a target position (i.e., the origin of coordinates), since the immediate rewards are greater the closer the helicopter hovers to the origin. Thus, the known space starts to expand (Figure 23 [b]) and, progressively, is concentrated at the origin of coordinates (Figure 23 [c] and [d]). With regard to the second matter, the probability of crashing is very low since, from the very beginning, the known space already appears concentrated at the origin and far from the error space (Figure 23 [b]). In other words, from the very beginning, all features of the known space (i.e., forward, sideways and downward velocities; x, y, and z coordinates; x, y and z angular-rates; and x, y and z quaternion) are very far from error space limits, decreasing the probability of visiting an error state.

In the previous experiments, the second step of PI-SRL has been performed using an initial case-base $B$ free of failures that is built into the first step of the algorithm. The following experiments show the performance of the second step of PI-SRL when different initial policies are used. Figure 24 (a) shows the performance of these policies used as initial policies. The continuous black line indicates the performance of the initial safe case-based policy $\pi_B$, with an average cumulative reward per episode of -85,130.11, used in the previous experiments prior to the execution of step two in the algorithm. The remaining lines in the Figure correspond to the performance of three different initializations of the case-base $B$ used in the new experiments, prior to the execution of step two of the algorithm. Using a very poor initial policy (dashed green lines) with which the helicopter crashed in nearly all of the episodes, the average cumulative reward per episode was calculated at -108,548.03. Using a different poor (albeit less poor) initial policy (continuous red lines) with which the helicopter crashed occasionally, the average cumulative reward per episode was -91,723.89. Finally, a near-optimal policy (dashed blue lines) whereby helicopter hovering is free of failures yields an average cumulative reward per episode of -13,940.1.

The Figure 24 (b) shows performance in the second step (improving the baseline behavior step) of PI-SRL, starting from a case-base $B$ corresponding to the very poor, poor and the near-optimal policies presented in Figure 24 (a). In Figure 24 (b), the dashed blue lines correspond to the use of a case-base $B$ containing the near-optimal policy, the continuous red lines correspond to the use of a case-base $B$ containing the poor policy and the dashed green lines correspond to the use of a case-base $B$ containing the very poor policy. All the experiments in the Figure have been conducted using a high level of risk in the domain ($\sigma = 9 \times 10^{-3}$). The graph indicates that with the use of a near-optimal policy for an initial policy and a high level of risk level, the case-base does not worsen performance which, in fact, appears to improve slightly. The second step of PI-SRL prevents the degradation of the initial performance of $B$, since no updates of cases in the case-base are made using bad episodes. In other words, the updates in $B$ are made with the cases gathered from episodes with a cumulative reward similar to that of the best episode found at a particular point and using a threshold $\Theta$ (whose value is set to 5% of the cumulative reward of the best episode). For example, if the cumulative reward of the best episode is -13,940.1, only the episodes with a cumulative reward higher than -14,637 are used to update the case-base (discarding the bad episodes or other episodes with failures). In this way, good sequences of experiences are provided to the updates, since it has been proven that good sequences of experiences can cause an adaptive agent to converge to a stable and useful policy, while bad sequences may cause an agent to converge to an unstable or poor policy (Wyatt, 1997). The solid red lines in Figure 24 (b) show that using a poor policy with failures as initial policy produces a higher number of failures than using an initial policy that is free of failures. However and despite the poor initialization, PI-SRL is nevertheless able to learn a near-optimal policy as well as when a policy free of failures is used to initialize $B$ (see lines corresponding to a high level of risk, $\sigma = 9 \times 10^{-3}$, in Figure 21 (a)). Finally, the dashed green lines in Figure 24 (b) show that the use of a very poor initial policy with many failures results in decreased performance and a higher number of failures produced, even though it is nevertheless able to learn better behavior. In this case, the algorithm falls into a local minimum, probably biased by the very poor initialization. In both cases with poor policies, the number of failures is higher at the beginning of the learning process and decreases as the learning process proceeds. While both the poor and very poor initial policies are very close to the error space, this is in stark contrast to the initial policy shown in Figure 23 which, from the very beginning, already appears concentrated at the origin, far from the error space.

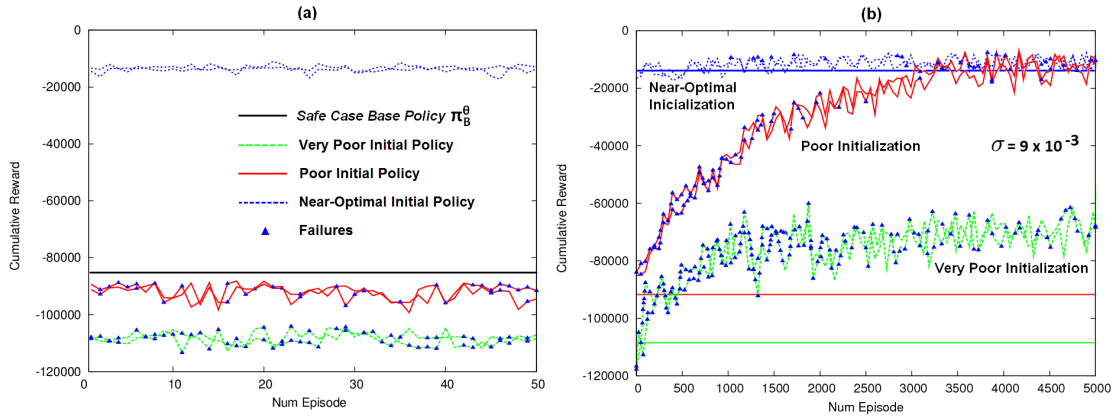

Figure 24: (a) The performance of different initial policies in the helicopter hovering task. (b) The performance of different executions of the second step of PI-SRL, each starting from a case-base $B$ containing a policy of three different types: very poor, poor and near-optimal.

As the learning process proceeds, the different policies are compressed away from the error space and the number of failures decreases.

## 4.4 SIMBA

Business simulators are powerful tools for improving management decision-making processes. An example of such a tool is the *SIMulator for Business Administration* (SIMBA) (Borrajo et al., 2010). SIMBA is a competitive simulator, since agents can compete against other agents through their management of different virtual companies. The simulator, the result of over twenty years of experience both with university students and business executives, emulates business realities using the same variables, relationships and events present in the business world. Its objective is to provide users with an integrated vision of a company, using basic techniques of business management, simplifying complexity and emphasizing the content and principles with the greatest educational value (Borrajo et al., 2010). In the experiments performed here, the learning agent competes against five hand-coded agents (Borrajo et al., 2010). Decision-making in SIMBA is an episodic task where decisions are made sequentially. To make a business decision, the state must be studied and 10 continuous decision variables (e.g., selling price, advertising expenses, etc.) must be set, followed by the study of a state composed of 12 continuous variables (e.g., material costs, financial expenses, economic productivity, etc.) (Borrajo et al., 2010). Each episode is composed of 52 steps, although it may prematurely if the company goes bankrupt (i.e., its losses are higher than 10% of its net assets).

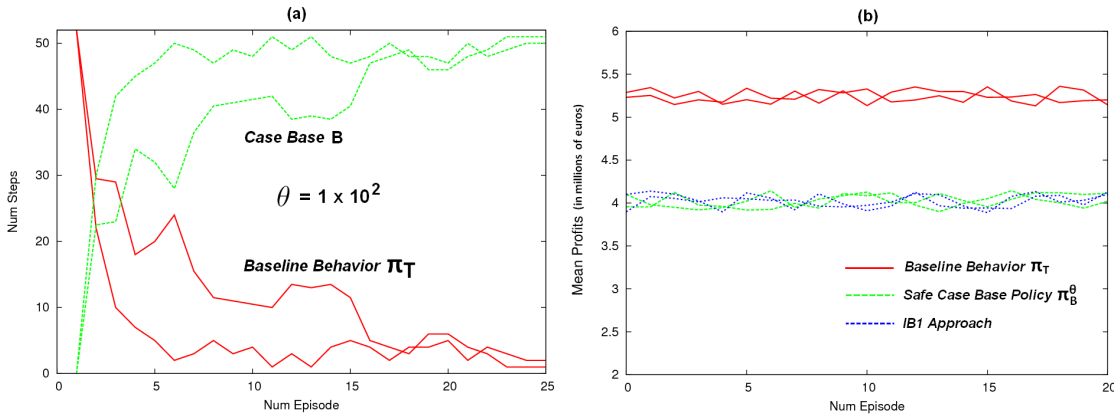

Figure 25: Modeling baseline behavior step in SIMBA Task: (a) Number of steps per trial executed by case-base $B$ and baseline behavior $\pi_T$. (b) Cumulative reward per trial by $\pi_T$, the learned safe case-based policy $\pi_B^\theta$ and an IBL approach.

Figure 25 (a) shows the evolution of the number of steps executed by the baseline behavior $\pi_T$ and the case-base $B$ during two learning processes performing the modeling baseline behavior step. $\theta$ and $\eta$ were computed following the procedure described in subsection 3.3 and have values of $1 \times 10^2$ and 513, respectively. In few episodes (approximately 25), the safe case-based policy $\pi_B^\theta$ is learned. Figure 25 (b) shows the performance of the previously-learned $\pi_B^\theta$, $\pi_T$ and the IB*1* approach. In this study, the mean profits per episode of $\pi_T$ are 5.24 million Euros, while those obtained for $\pi_B^\theta$ are 4.02 million Euros. In the IB*1* approach, all cases generated using the baseline behavior $\pi_T$ during 25 episodes are stored. The experiments demonstrate that in SIMBA, in contrast with the previous domains, storing all cases is sufficient for obtaining a safe policy with a performance similar to that using the modeling baseline behavior step (with mean profits per episode of 3.98 million Euros). Once the safe case-based policy $\pi_B^\theta$ is learned, we execute the improving the learned baseline behavior step.

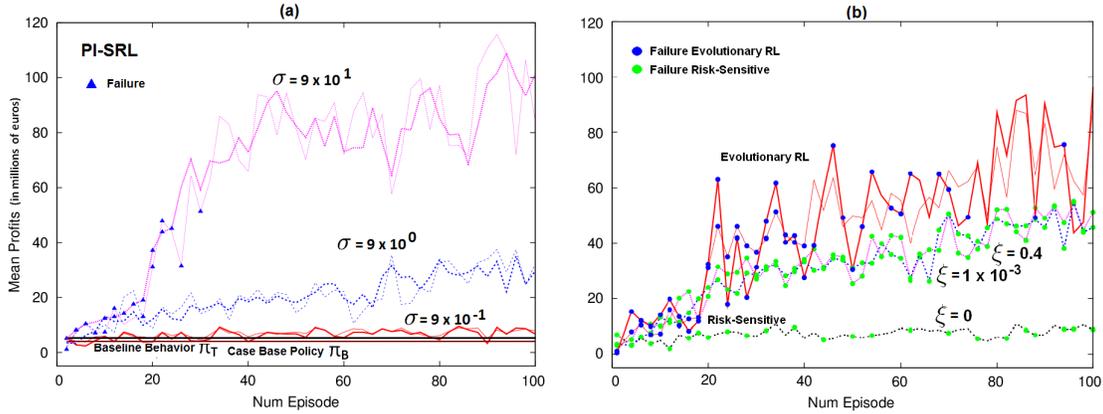

Figure 26: Improving the learned baseline behavior step in SIMBA task: (a) The mean profits per episode for different risk configurations obtained by the PI-SRL agent against five hand-coded agents. (b) The mean profits per episode obtained by the evolutionary and risk-sensitive RL agent against five hand-coded agents. In each cases, any episode ending in failure (bankruptcy) is noted.

Similar to the findings in earlier tasks, Figure 26 (a) indicates that while low and medium levels of risk do not produce bankruptcies, performance is nevertheless weak. The highest level of risk produces a near-optimal policy with a low number number of failures. By contrast, Figure 26 (b) presents the results for the evolutionary and risk-sensitive RL approaches, with the former being clearly that which yields the highest number of failures. In the risk-sensitive case, the number of bankruptcies in all cases is insufficient for learning the risk function $\rho$. The comparative results in Figure 26 show that PI-SRL with $\sigma = 9 \times 10^1$ obtains better policies and less failures than the evolutionary or risk-sensitive RL approaches.

Figure 27 shows a graphical representation of the different solutions in this domain. It shows the mean number of failures and cumulative reward for the different approaches over 100 episodes, with data computed from 10 independent executions of each approach. In the Figure, red circles correspond to the PI-SRL algorithm, black triangles correspond to the risk-sensitive approach and the blue square corresponds to the evolutionary RL approach. Figure 27 also shows the performance for two additional risk levels, very high ($\sigma = 9 \times 10^2$) and very low ($\sigma = 0$), with respect the Figure 26. The experiments in Figure 27 demonstrate that PI-SRL with a high level of risk ($\sigma = 9 \times 10^1$) obtains the highest cumulative reward, 6693.58. Additionally, PI-SRL with a very low level of risk ($\sigma = 0$), a low level of risk ($\sigma = 9 \times 10^{-1}$) and a medium level of risk ($\sigma = 9 \times 10^0$) are the approaches with the lowest mean number of failures, 0.0. However, PI-SRL with a medium level of risk is preferred inasmuch as its performance is superior in terms of cumulative reward. PI-SRL with a very high level risk ($\sigma = 9 \times 10^2$) increases the number of failures and obtains a lower cumulative reward when compared to PI-SRL with a high level of risk. Using the Pareto comparison criterion, PI-SRL with a high level of risk strictly dominates all other solutions (PI-SRL $\sigma = 9 \times 10^1 \succ$ risk-sensitive and PI-SRL $\sigma = 9 \times 10^1 \succ$ evolutionary RL), while the approach is not strictly dominated by any other solution.

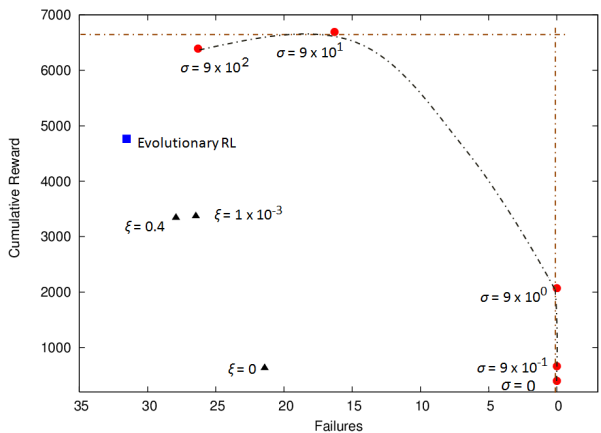

Figure 27: Mean number of failures (company bankruptcies) and the cumulative reward over 100 episodes for each approach to the SIMBA task. The means have been computed from 10 different executions.

Due to the difficulty of representing the high-dimensional state and action space of the SIMBA domain, no graphs are provided with the evolution of the known space.

## 5. Related Work

Reinforcement learning (RL) and case-based reasoning (CBR) techniques have been combined in the literature in different ways. In the work of Bianchi et al. (2009), a new approach is presented permitting the use of cases as heuristics to speed up RL algorithms. Additionally, Sharma et al. (2007) use a combination of CBR and RL (called CARL) to achieve transfer while playing against the Game AI across a variety of scenarios in MadRTS TM, a commercial Real Time Strategy game. CBR has also been used for state value function approximation in RL (Gabel & Riedmiller, 2005). However, the present study is, to our knowledge, the first time that CBR and RL have been used in conjunction for safe exploration in dangerous domains. In the field of *safe reinforcement learning*, three principal trends can be observed: (i) approaches based on return and its variance, (ii) risk-sensitive approaches based on the definition of error states and (iii) approaches using teachers.

### 5.1 Approaches Based on the Return and its Variance

In the literature, it has long been known that the optimal policy and the optimal expected return of an MDP are quite sensitive to parameter variations (even an optimal policy may perform badly in some cases due to the stochastic nature of the problem). To mitigate this problem, the agent can try to maximize the return associated with the worst-case scenario, even though the case may be highly unlikely. Thus, in this trend, the risk refers to the worst outcomes of the return $R = \sum_{t=0}^{\infty} \gamma^t r_t$ or its variance. An example of such an approach is *worst-case control* where the worst possible outcome of $R$ is to be optimized (Coraluppi & Marcus, 1999; Heger, 1994). In worst case control strategies, the optimality criterion is exclusively focused on risk-avoiding policies. A policy is considered to be optimal if its worst-case return is superior. The approach, however, is too restrictive inasmuch as it takes very rare scenarios fully into account.

The $\alpha - value$ of the return $\hat{m}_\alpha$ introduced by Heger (1994) can be seen as an extension of the worst case control of MDPs. This concept establishes that the returns $R < \hat{m}_\alpha$ of a policy that occur with a probability lower than $\alpha$ are neglected. The algorithm is less pessimistic than pure worst case control, given that extremely rare scenarios have no effect on the policy. In the work of Heger et al., the idea of weighting return and risk, namely the *expected value-variance criterion*, is also introduced.

In risk-sensitive control based on the use of *exponential utility functions*, the return $R$ is transformed to reflect a subjective measure of utility. Instead of maximizing the expected value of $R$, the objective here is to maximize $U = \beta^{-1} log E(e^{\beta R})$, where $\beta$ is a parameter and $R$ is the usual return. It can be shown that, depending on the parameter $\beta$, policies with a high variance $V(R)$ are penalized ($\beta < 0$) or enforced ($\beta > 0$). Instead, Neuneier and Mihatsch (2002) consider the *worst-case-outcomes* of a policy, (i.e., risk related to the variability of the return). In the study, the authors demonstrate that the learning algorithm interpolates between *risk-neutral* and the worst-case criterion and has the same limiting behavior as exponential utility functions. It should be noted that these approaches based on the variability of the return or its worst possible outcomes are not suited for problems where a policy with a small variance can produce a large risk (Geibel & Wysotzki, 2005). Our view of risk in the present study, however, is not concerned with the variance of the return or its worst possible outcome, but instead with the fact that processes generally possess unsafe states that should be avoided. Consequently, we address a different class of problems than those dealt with by approaches focusing on the variability of the return.

## 5.2 Risk-sensitive Approaches based on Error States.

In this second trend of approaches, the concept of risk is based on the definition of *error states* or fatal transitions. Thus, Geibel et al. (2005) , for instance, establish the risk function as the probability of entering in an error state. Instead, Hans et al (2008) consider a transition to be fatal if the corresponding reward is less than a given threshold $\tau$. In the first case and as demonstrated in Section 4, $\rho$ is learned by TD methods which require that error states (i.e., car collisions, pole-balancing disequilibrium, helicopter crashes and company bankruptcies) be visited repeatedly in order to approximate the risk function and, subsequently, avoid dangerous situations. In the second case, the concept of risk is again joined with that of reward. Moreover, the above mentioned studies either (i) assume that the system dynamics are known, (ii) tolerate undesirable states during exploration or, in contrast with our paper, (iii) do not deal with problems with high-dimensional and continuous state-action spaces. Regarding the latter, while Geibel et al. write that their approach can also be extended to continuous action sets (e.g., by using an actor-critic method), they do not give any more details on how this may be done with entirely continuous problems. In Section 4, we present an approach that solves the problem.

## 5.3 Approaches Using Teachers

The last trend in the approaches is based on the use of teachers in three different ways: (i) to bootstrap the learning algorithm (i.e., as an initialization procedure), (ii) to derive a policy from a finite demonstration set and (iii) to guide the exploration process.

### 5.3.1 Bootstrapping the Learning Algorithm

In the work of Driessens and Sžeroski (2004), a bootstraping procedure is used for relational RL in which a finite set of demonstrations are recorded from a human expert and later presented to a regression algorithm. This allows the regression algorithm to build a partial Q-function which can later be used to guide further exploration of the state space using a Boltzmann exploration strategy. Smart and Kaelbling (2000) also use examples, training runs to bootstrap the Q-learning approach for their HEDGER algorithm. The initial knowledge bootstrapped into the Q-learning approach allows the agent to learn more effectively and helps reduce the time spent with random actions. Teacher behaviors are also used as a form of *population seeding* in neuroevolution approaches (Yao, 1999; Siebel & Sommer, 2007). Evolutionary methods are used to optimize the weights of neural networks, but starting from a prototype network whose weights correspond to a teacher (or baseline policy). Using this technique, RL Competition helicopter hovering task winners Martin et al. (2009) developed an evolutionary RL algorithm in which several teachers are provided in the initial population. The algorithm restricts crossover and mutation operators, allowing only slight changes to the policies given by the teachers. Consequently, the rapid convergence of the algorithm to a near-optimal policy is ensured, as is the indirect minimization of damage to the agent. However, the teachers included in the initial population resulting from an ad-hoc training regimen conducted before the competition. Consequently, the proposed approach seems somewhat ad-hoc and not easily generalizable to arbitrary RL problems. In the work of Koppejan et al. (2009, 2011), neural networks are also evolved, beginning with one whose weights corresponds to teacher behavior. While this approach has been proven advantageous in numerous applications of evolutionary methods (Hernández-Díaz et al., 2008; Koppejan & Whiteson, 2009), Koppejan's algorithm nevertheless also seems somewhat ad-hoc and designed for a specialized set of environments.

### 5.3.2 Deriving a Policy from a Finite Set of Demonstrations

All approaches falling under this category are framed according to the field of Learning from Demonstration (LfD) (Argall et al., 2009). Highlighting the study by Abbeel et al. (2010) based on apprenticeship learning, the approach is composed of three distinct steps. In the first, a teacher demonstrates the task to be learned and the state-action trajectories of the teacher's demonstration are recorded. In the second step, all state-action trajectories seen to that point are used to learn a dynamics model for the system. For this model, a (near-)optimal policy is to be found using any reinforcement learning (RL) algorithm. Finally, the policy obtained should be tested by running it on the real system. In the work of Tang et al. (2010), an algorithm based on apprenticeship learning is also presented for automatically-generating trajectories for difficult control tasks. The proposal is based on the learning of parameterized versions of desired maneuvers from multiple expert demonstrations. Despite each approach's potential strengths and general interest, all are inherently linked to the information provided in the demonstration dataset. As a result, learner performance is heavily limited by the quality of the teacher's demonstrations.

### 5.3.3 Guiding the Exploration Process

Driessens and Sžeroski (2004), in the context of relational RL, also use a given teacher's policy, rather than a policy derived from the current Q-function hypothesis (which is not informative in the early learning stages), for the selection of actions. In this approach, episodes performed by a teacher are interleaved with normal exploration episodes. This mixture of teacher and normal exploration make it easier for the regression algorithm to distinguish between beneficial and poor actions. In the context of LfD, there are other approaches which include teacher advice (Argall et al., 2009). This advice is used to improve learner performance, offering information beyond that which is provided by a demonstration dataset. In this approach, following an initial task demonstration by the teacher, the agent directly requests additional demonstration from the teacher in very different states from those previously demonstrated or in states in which a single action cannot be selected with certainty (Chernova & Veloso, 2007, 2008).

In all works mentioned for this trend, no explicit definition of risk is ever given.

## 6. Conclusions

In this work, PI-SRL, an algorithm for policy improvement through safe reinforcement learning in high-risk tasks, is described. The main contributions of this algorithm are the definitions of a novel case-based risk function and a baseline behavior for the safe exploration of the state-action space. The use of the case-based risk function presented is possible inasmuch as the policy is stored as a case-base. This represents a clear advantage over other approaches, e.g., evolutionary RL (Martín H. & Lope, 2009; Koppejan & Whiteson, 2011) where the extraction of knowledge about the known space by the agent is impossible using the weights of the neural-networks. Additionally, a completely different notion of risk from others found in the literature is presented. According to our notion, risk is independent of the variance of the return and the reward function, and does not require the identification of error states or the learning of risk functions. Rather, the concept of risk described in this paper is based on the distance between the known and unknown space and, therefore, is a domain-independent parameter (in this sense, our proposal allows for the application of a parameter-setting method as described in subsection 3.3). While Koppejan et al. (2011) also use a function to identify dangerous states, in contrast with our approach, the definition of their function requires strong previous knowledge of the domain. Furthermore, most of the approaches to risk found in the literature only tackle problems that are not entirely continuous (Geibel & Wysotzki, 2005) or that only report results on one continuous domain (Koppejan & Whiteson, 2011). Consequently, it is difficult to know for certain if these approaches from the literature generalize easily to arbitrary domains.

This paper presents the PI-SRL algorithm in great detail and demonstrates its effectiveness in four entirely different continuous domains: the car parking problem, pole-balancing, helicopter hovering and business management (SIMBA). The experiments presented in this paper demonstrate different characteristics about the learning capabilities of the PI-SRL algorithm.

*(i) PI-SRL obtains higher quality solutions.* The experiments in Section 4 demonstrate that, save in the helicopter hovering task, PI-SRL obtains in all cases the best cumulative reward per episode and the least number of failures. Additionally, using the Pareto comparison criterion it can be said that, save the very high risk configuration in the car parking problem, our approach is not strictly dominated by any other approach.

*(ii) PI-SRL adjusts the initial known space to safe and better policies.* The initial known space resulting from the first step of PI-SRL, modeling baseline behavior, is adjusted and improved in the second step of the algorithm, improving the learned baseline behavior. Additionally, the experiments demonstrate that the adjustment process can compress the known space away from the error space (e.g., pole-balancing domain, subsection 4.2, and helicopter hovering domain, subsection 4.3) or, on other occasions, can require the known space to move closer to the error space (e.g., car parking problem, subsection 4.1) in the event that better policies are be found there.

*(iii) PI-SRL works well in domains with differently structured state-action spaces and where the value function can vary sharply.* Although the car parking problem, the pole-balancing domain, the helicopter hovering task and the business simulator all represent very differently structured problems, experiments in the study nevertheless demonstrate that PI-SRL performs well in each. Furthermore, even in such domains as the car parking problem in which the value function varies sharply due to the presence of an obstacle, experimental results demonstrate that PI-SRL can nevertheless successfully handle this difficulty. However, it is impossible to avoid all failures if the "known space" edge is the same as the edge to error states the algorithm would often 'explore' into error states.

*(iv) The number of failures depends on the distance between the known space and the error space.* The experiments in the pole-balancing and helicopter hovering domains demonstrate that the number of failures depends on how close the known space is to the error space. Due to the structure of these domains, the improving the learned baseline behavior step in the algorithm tends to concentrate the known space at the origin of coordinates away from the error space. The greater the distance between the known space and the error space, the lower the number of failures. Additionally, in helicopter hovering, the known space is, from the beginning, far from the error space (consequently, the number of failures is also low from the beginning). Therefore, the initial distribution of the known space learned from the baseline policy $\pi_T$ later influences the number of failures obtained by the second step of PI-SRL.

*(v) PI-SRL is completely safe if only the first step of the algorithm is executed.* However, by proceeding only in this way, algorithm performance would be heavily limited by the capabilities of the baseline behavior. If learner performance is to be improved beyond the performance of this baseline behavior, the subsequent exploratory process from the second step of PI-SRL must be carried out. Since complete knowledge of the domain and its dynamic is not possessed, however, it is also inevitable that, during this exploratory process, unknown regions of the state space will be visited where the agent may reach error states.

*(vi) The risk parameter allows the user to configure the level of risk assumed.* In our algorithm, the user can gradually increase the value of the risk parameter $\sigma$ in order to obtain better policies, but also assuming a greater likelihood of damage in the learning system.

*(vii) PI-SRL performs successfully even when a poor initial policy with failures is used.* The experiments in Figure 24 from the helicopter hovering domain demonstrate that PI-SRL is able to learn a near-optimal policy despite poor initialization, just as it can when a policy free of failures is used to initialize the case-base $B$. However, the Figure also shows that if a very poor initial policy with many failures is used, PI-SRL decreases in performance and produces a higher number of failures, although some better behavior is still learnt. In this case, the algorithm falls into a local minimum, likely biased by the very poor initialization.

In what follows, the applicability of the method is discussed, allowing the reader to more clearly understand the scenarios in which the proposed PI-SRL approach may be applicable. This applicability is restricted to domains having the following characteristics.

*(i) It is mandatory that the scenario satisfy the two assumptions described in Section 2.* According to the first assumption, nearby states in the domain must necessarily have similar actions. According to the other, similar actions in similar states should produce similar effects. This fact that similar actions lead to similar states assumes some degree of smoothness in the dynamic behavior of the system which, in certain environments, may not hold. However, as we clearly explain in Section 2, we consider both assumptions to be logical assumptions derived from generalization principles in the RL literature (Kaelbling et al., 1996; Jiang, 2004).

*(ii) The applicability of the method is limited by the size of the case-base $B$ required to mimic the baseline behavior.* It is not possible to apply the proposed approach to tasks when, in the first step of the PI-SRL algorithm, modeling baseline behavior, a prohibitively large number of cases are required to properly mimic complex baseline behaviors. In this case, the threshold $\theta$ can be increased to further restrict the addition of new cases to the case-base. However, this increase may adversely affect the final performance of the algorithm. Nevertheless, the experiments performed in Section 4 demonstrate that relatively simple baseline behaviors are mimicked almost perfectly using a manageable number of cases.

*(iii) The PI-SRL algorithm requires the presence of a baseline behavior.* The proposed method requires the presence of a baseline behavior that safely demonstrates the task to be learned. This baseline behavior can be conducted by a human teacher or a hand-coded agent. It is important to note, nevertheless, that the presence of such a baseline behavior is not guaranteed in all domains.

Finally, a logical continuation of the present study would take into account the automatic graduation of the risk parameter along the learning process. For example, it would be particularly interesting to exploit the fact that the known space is far away from the error space in order to increase the risk parameter or, on the contrary, to reduce it when it is close. Other future work aims to deploy the algorithm in real environments, inasmuch as the uncertainty of the real environments presents the biggest challenge to autonomous robots. Autonomous robotic controllers must deal with a large number of factors such as the robotic mechanical system and electrical characteristics, as well as environmental complexity. However, the use of the PI-SRL algorithm (or other risk-sensitive approaches) for learning processes in real environments could reduce the amount of damage incurred and, consequently, allow the lifespan of the robots to be extended. It might be worthwhile add a mechanism to the algorithm to detect when a known state can lead directly to an error state. All such problems are currently being investigated.

## Acknowledgments

This study has been partially supported by Spanish MICIIN projects TIN2008-06701-C03-03, TRA2009-0080 and CCG10-UC3M/TIC-5597. We offer our gratitude and special thanks to Raquel Fuentetaja Pizán, Assistant Professor at Universidad Carlos III de Madrid in the Planning & Learning Group (PLG), for her generous and invaluable comments during the revision of this paper. We would also like to thank to José Antonio Martín, Assistant Professor at Universidad Complutense de Madrid, for his invaluable comments regarding his evolutionary RL algorithm.